\documentclass[11pt]{article}

\PassOptionsToPackage{table}{xcolor}
\usepackage[final]{acl}
\usepackage{times}
\usepackage{latexsym}
\usepackage[T1]{fontenc}
\usepackage[utf8]{inputenc}
\usepackage{microtype}
\usepackage{inconsolata}
\usepackage{graphicx}
\usepackage{subcaption}
\usepackage{booktabs}
\usepackage{longtable}
\usepackage{tabularx}
\usepackage{tikz}
\usetikzlibrary{positioning,calc,fit}

\usepackage{amsmath}
\usepackage{amssymb}
\usepackage{mathtools}
\usepackage{amsthm}

\usepackage[capitalize,noabbrev]{cleveref}

\theoremstyle{plain}

\theoremstyle{definition}

\theoremstyle{remark}

\usepackage[textsize=tiny]{todonotes}

\title{Efficient GUI Agents: A Systems Survey of Observation,\\
Memory, Action, and Runtime Optimization}

\author{
  \textbf{Bizhe Bai\textsuperscript{1,2,\textdagger}},
  \textbf{Jiakang Yuan\textsuperscript{1,\textdagger}},
  \textbf{Hongming Wu\textsuperscript{1}},
  \textbf{Xinyue Wang\textsuperscript{1}},
\\
  \textbf{Jie Ren\textsuperscript{1}},
  \textbf{Siyao Chen\textsuperscript{1}},
  \textbf{Yuchen Ya\textsuperscript{1}},
  \textbf{Fan Bai\textsuperscript{3}},
\\
  \textbf{Pai Peng\textsuperscript{3}},
  \textbf{Huafeng Qin\textsuperscript{4}},
  \textbf{Tao Chen\textsuperscript{1,2}}
\\[0.8em]
  \textsuperscript{1}College of Future Information Technology,
  Fudan University, Shanghai, China
\\
  \textsuperscript{2}Shanghai Innovation Institute,
  Shanghai, China
\\
  \textsuperscript{3}Independent Researcher
\\
  \textsuperscript{4}Chongqing Technology and Business University  \\[0.5em]
}
\newcolumntype{Y}{>{\raggedright\arraybackslash}X}
\newcolumntype{L}[1]{>{\raggedright\arraybackslash}p{#1}}
\definecolor{ghblack}{HTML}{111111}
\definecolor{revisiongreen}{HTML}{008000}
\newcommand{\codelink}[1]{\href{#1}{\begingroup\setlength{\fboxsep}{1pt}\colorbox{ghblack}{\textcolor{white}{\scriptsize GitHub}}\endgroup}}
\newcommand{\na}{--}
\newcommand{\revgreen}[1]{#1}
\newcommand{\tablegroup}[1]{\addlinespace[2pt]\multicolumn{7}{@{}l}{\textbf{#1}}\\[-0.2em]}
\begin{document}

\maketitle

\begin{abstract}
  GUI agents increasingly operate across websites, mobile apps, and desktop environments, yet the field still reports progress primarily through task success. We argue that practical deployment depends equally on efficiency: how much context, computation, action budget, and runtime overhead an agent consumes while succeeding. This survey studies efficient GUI agents through an end-to-end systems lens that preserves the current technical axes of observation efficiency, context and memory efficiency, action efficiency, and planner-side/system efficiency. For each subsection, we expand the seed literature through targeted search plus backward and forward citation chaining, then synthesize the dominant mechanisms, reported efficiency signals, and new overheads they introduce. Across the literature, recent progress converges on a small set of recurring ideas: selective reading instead of full-context ingestion, global-to-local visual allocation, recoverable memory rather than raw history replay, verification-aware control, and hybrid runtimes that can switch between GUI and non-GUI execution. We conclude by identifying the main open problems, including honest accounting of verifier cost, cross-benchmark comparability, and co-design of observation, memory, and execution layers under real latency and privacy constraints.
\end{abstract}

\section{Introduction}
\label{sec:introduction}
Large language models (LLMs) and vision-language models (VLMs) have moved GUI automation toward open-ended computer use: modern agents can interpret natural-language instructions, inspect live graphical interfaces, and carry out multi-step tasks across websites, mobile applications, and desktop operating systems~\cite{efficientagentssurvey,computerusesurvey,guiagentssurvey}. This progress is reflected in interactive benchmarks such as Mind2Web, WebArena, VisualWebArena, BrowserGym, AndroidWorld, OSWorld, and Windows Agent Arena, which evaluate agents on web navigation, visually grounded web interaction, mobile-app control, and open-ended operating-system tasks~\cite{mind2web,webarena,visualwebarena,browsergym,rawles2024androidworld,osworld,windowsagentarena}. Collectively, these benchmarks show that recent GUI agents are becoming capable of increasingly realistic computer-use tasks, from navigating complex websites and operating mobile apps to completing multi-step workflows in desktop environments.

At the same time, these capabilities often come with substantial interaction and runtime costs. A GUI agent may eventually complete a task, but only after many observation--reasoning--action cycles, repeated model calls, redundant interface operations, or long waits between steps. OSWorld-Human, for example, shows that leading computer-use agents on OSWorld still take substantially more steps than human-derived trajectories and that planning, judging, and reflection dominate end-to-end latency~\cite{abhyankar2025osworldhuman}. This matters in practice because GUI agents interact with the same visible interfaces that users depend on: an agent that occupies the screen for too long, performs unnecessary actions, or requires users to wait through slow intermediate steps can be difficult to deploy even when its final answer is correct. Recent efficiency-aware studies have therefore begun to profile step counts, visual-token and cache costs, planning latency, action abstraction, and end-to-end runtime overhead in GUI-agent execution~\cite{huang2025guikv,zhong2026actionengine,abhyankar2026osworldhumanbenchmarkingefficiencycomputeruse}. These findings expose a limitation of success-only evaluation: an agent that eventually completes a task may still be too slow, too costly, too context-heavy, or too interaction-inefficient for practical deployment.


Following recent surveys on efficient agents, computer-use agents, and GUI agents~\cite{efficientagentssurvey,computerusesurvey,guiagentssurvey}, we distinguish among three related notions:

\noindent\fbox{\parbox{0.97\linewidth}{
\textbf{LLM.} A foundation model trained on large-scale language data, often used as the reasoning and natural-language-processing backbone of agentic systems.\\[0.25em]
\textbf{Agent.} A goal-directed system that repeatedly observes an environment, updates its internal state, and selects actions to make progress toward a task objective.\\[0.25em]
\textbf{GUI agent.} A specific type of computer-use agent whose primary perception and actuation channels are graphical user interfaces rather than text-only APIs or command lines.\\[0.25em]
}}

In this survey, we use \emph{agent} in the above closed-loop sense. A \emph{GUI agent} is therefore not simply an LLM prompted with screenshots: it is a computer-use agent whose observations and actions are tied to screenshots, DOM or HTML structures, accessibility trees, focused elements, window metadata, and interface-level operations such as clicking, typing, scrolling, dragging, invoking shortcuts, or switching applications~\cite{mind2web,webarena,rawles2024androidworld,osworld,guiagentssurvey,computerusesurvey}. Unlike text-only tool agents, GUI agents must solve visual grounding, partial observability, long-horizon state tracking, and execution under noisy or incomplete interface representations~\cite{visualwebarena,osworld,guiagentssurvey,computerusesurvey}.
\textbf{Why is efficiency particularly critical for GUI agents?} GUI interaction is intrinsically multimodal and observation-intensive. On the web, agents may need to parse long DOM or accessibility trees whose redundant structure inflates context length and reasoning cost \cite{abuelsaad2024agente,schiepanski2025beyondpixels,kerboua2025focusagent,zhang2025prune4web}. In screenshot-centric settings, the agent must interpret visually dense screens, small targets, and weakly structured accessibility metadata, which makes grounding both computationally expensive and error-prone \cite{cheng2024seeclick,you2024ferretui,gou2024uground,osworld,windowsagentarena}. More broadly, literature on efficient multimodal-LLM indicates that multimodal capability is often constrained by training cost, inference latency, memory pressure, and repeated visual computation \cite{efficientmllmsurvey}. These issues become sharper in GUI environments because perception, reasoning, and action are interleaved at every step rather than amortized over a single query. GUI tasks are also long-horizon and error-sensitive: agents must recover from failed actions, interface changes, and lengthy trajectories, and human profiling already shows large step and latency gaps in realistic computer-use tasks \cite{agashe2024agents,abhyankar2025osworldhuman,kang2026longhorizonui,rawles2024androidworld}. Efficiency also shapes deployability and privacy, since practical systems may need to minimize exposed UI content, cloud traffic, and runtime overhead \cite{efficientmllmsurvey,fan2025core,wang2026guiguard}.


This survey makes two contributions. First, we organize the literature with an efficiency-centered taxonomy spanning observation, context and memory, action, and planner-side/system optimization. Second, we synthesize the main cross-layer trade-offs and open challenges, including cases where apparent savings are offset by new parser, verifier, retriever, or orchestration costs.

\section{Preliminaries}
\label{sec:background_problem}

This section fixes the terminology and the system view used in the rest of the survey. The goal is not to formalize every implementation detail, but to provide a compact description of what a GUI agent is, how it works, and where efficiency enters the loop.

\subsection{From Agents to GUI Agents}
\label{subsec:agent_formulation}

From the perspective of this survey, GUI agents are a specific class of agentic systems: they receive a user goal, observe a graphical interface, reason about the next step, and execute an interface-level action \cite{efficientagentssurvey,computerusesurvey}. A compact way to write this loop is
\begin{equation}
a_t \sim \pi(g, o_t, m_t),
\end{equation}
where $g$ is the user goal, $o_t$ is the current GUI observation, $m_t$ is the retained context or memory, and $a_t$ is the next action. We use this formulation only to fix notation. In practice, $o_t$ may include screenshots, DOM or HTML, accessibility trees, or related interface metadata, while $a_t$ may include clicking, typing, scrolling, dragging, hotkeys, or app switching \cite{mind2web,webarena,rawles2024androidworld,osworld}. 
A more detailed discussion of  GUI-agent is  stated in  Appendix \ref{sec:gui-agent-form}

\subsection{How GUI Agents implement and What Should Be Efficient}
\label{subsec:gui_agent_workflow}

Although implementations differ, most GUI agents follow the same functional pipeline. They first construct an actionable representation of the current interface from screenshots, DOM or HTML, accessibility trees, or hybrid parsing outputs \cite{mind2web,webarena,cheng2024seeclick,gou2024uground}. They then retain task-relevant context through short-horizon history or explicit memory structures, reason about the next subgoal or action, ground that decision to an executable element or coordinate, execute the action, and verify whether the intended effect occurred \cite{agashe2024agents,osworld,lee2025verisafe,chen2025guishepherd,kang2026longhorizonui}. Some systems collapse these stages into a single multimodal model, while others expose them as separate modules; the workflow remains the same.

\begin{figure}[t]
    \centering
    \includegraphics[width=\linewidth]{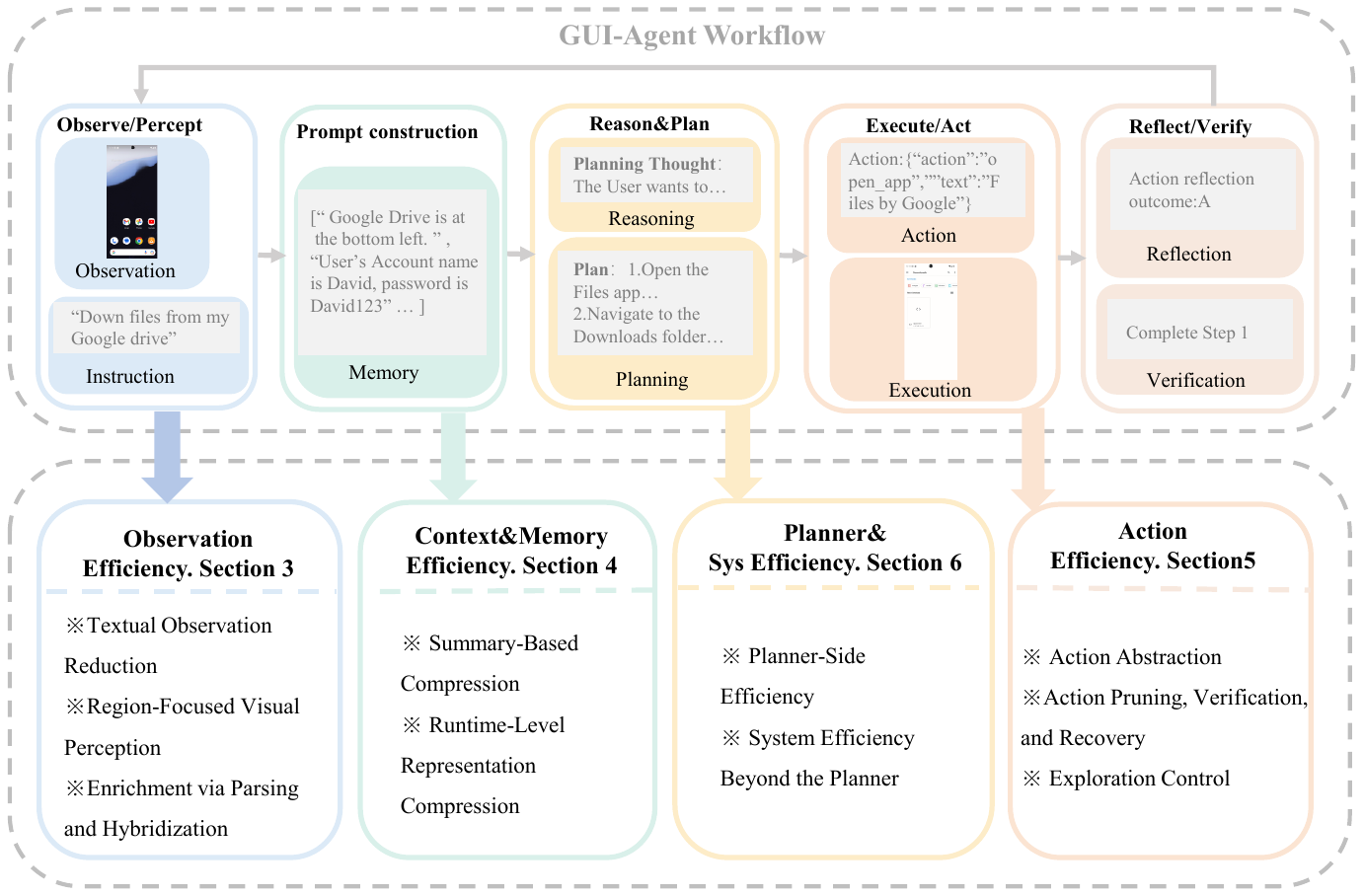}
    \caption{A GUI-agent system can be viewed as a loop of observation/perception, context and memory, planning/reasoning, grounding/action execution, and verification/feedback. The four taxonomy axes of this survey are shown explicitly on the corresponding stages: Observation Efficiency, Context and Memory Efficiency, Action Efficiency, and Planner-Side and System Efficiency.}
    \label{fig:gui_agent_system_overview}
\end{figure}
This workflow also clarifies what this survey means by efficiency. \emph{Observation efficiency} concerns how the current interface is represented without overwhelming the model with redundant text or visual tokens \cite{abuelsaad2024agente,schiepanski2025beyondpixels,lin2025showui}. \emph{Context and memory efficiency} concerns how historical information is compressed, retained, or retrieved across steps \cite{liu2025guirise,liu2025palui}. \emph{Action efficiency} concerns how economically the agent reaches the goal through abstraction, pruning, verification, and recovery \cite{zhang2025prune4web,lee2025verisafe,kang2026longhorizonui}. \emph{Planner-side and system efficiency} concerns reasoning depth, orchestration overhead, backend routing, and end-to-end runtime beyond the next-action predictor itself \cite{efficientagentssurvey,fan2025core,wang2026guiguard}.

\subsection{Efficiency Metrics for GUI Agents}
\label{subsec:efficiency_metrics}
Objective evaluation should separate \emph{effectiveness} from \emph{efficiency}. Effectiveness metrics measure task quality; efficiency metrics measure the resources spent to obtain that quality. We use the following vocabulary:
\begin{enumerate}
    \item \textbf{Latency metrics}: TTFT is time from request dispatch to the first generated token; TPOT is average time per generated token after the first token; inference latency is total model-call time; per-step latency is one observe--plan--act--verify cycle; end-to-end latency is time from task start to success or failure.
    \item \textbf{Observation metrics}: perception calls count screenshot, crop, OCR, VLM, DOM, HTML, or AxTree parsing invocations; DOM/HTML/AxTree length measures serialized interface size; image crops count localized visual inputs.
    \item \textbf{Context and memory metrics}: history length, retrieved memory size, KV-cache size, and memory footprint measure retained context, retrieved state, and serving memory cost.
\end{enumerate}

\subsection{Organization of This Survey}
\label{subsec:paper_organization}

The remainder of this survey is organized as follows. Section 3 reviews Observation Efficiency. Section 4 reviews Context and Memory Efficiency. Section 5 reviews Action Efficiency. Section 6 discusses Planner-Side and System Efficiency. Section 7 outlines Open Challenges and Future Directions.  For readability, the tree-structured taxonomy is collected in Figure~\ref{fig:taxonomy_overview}, while the section-wise paper summary tables are collected in the Appendix \ref{sec:section-wise-table}.
\begin{figure*}[h]
\centering
\resizebox{0.92\textwidth}{!}{\providecommand{\papercite}[2]{\mbox{#1~[\citenum{#2}]}}

\begin{tikzpicture}[x=1cm,y=1cm]
  \definecolor{rootfill}{HTML}{F0F0F0}
  \definecolor{obsmajor}{HTML}{DCEAF7}
  \definecolor{obsminor}{HTML}{EDF5FC}
  \definecolor{memmajor}{HTML}{D9F0EA}
  \definecolor{memminor}{HTML}{EDF8F5}
  \definecolor{actmajor}{HTML}{FBE4D2}
  \definecolor{actminor}{HTML}{FEF2E8}
  \definecolor{sysmajor}{HTML}{F7E8DE}
  \definecolor{sysminor}{HTML}{FDF5EF}

  \tikzset{
    connector/.style={
      draw=black!72,
      line width=0.58pt,
      rounded corners=1.5pt
    },
    root box/.style={
      rounded corners=3pt,
      draw=black!75,
      line width=0.75pt,
      fill=rootfill,
      text width=2.15cm,
      minimum height=0.95cm,
      align=center,
      inner sep=5pt,
      font=\bfseries\small
    },
    major topic box/.style={
      rounded corners=3pt,
      draw=black!70,
      line width=0.72pt,
      text width=2.55cm,
      minimum height=0.95cm,
      align=center,
      inner sep=5pt,
      font=\bfseries\small
    },
    subtopic box/.style={
      rounded corners=2.5pt,
      draw=black!65,
      line width=0.62pt,
      text width=3.55cm,
      minimum height=0.92cm,
      align=center,
      inner sep=4pt,
      font=\bfseries\footnotesize
    },
    paper list box/.style={
      rounded corners=2.5pt,
      draw=black!55,
      line width=0.52pt,
      text width=5.95cm,
      minimum height=0.92cm,
      align=left,
      inner sep=4pt,
      font=\scriptsize
    }
  }

  \node[root box] (rootAgent) at (0.2,2.96)
    {\textbf{Efficient}\\\textbf{GUI Agent}};

  \node[major topic box,fill=obsmajor] (majorObs) at (3.25,9.60)
    {\textbf{Observation}\\\textbf{Efficiency}};
  \node[major topic box,fill=memmajor] (majorMem) at (3.25,5.18)
    {\textbf{Context \& Memory}\\\textbf{Efficiency}};
  \node[major topic box,fill=actmajor] (majorAct) at (3.25,0.75)
    {\textbf{Action}\\\textbf{Efficiency}};
  \node[major topic box,fill=sysmajor] (majorSys) at (3.25,-3.68)
    {\textbf{Planner-Side \&}\\\textbf{System Efficiency}};

  \node[subtopic box,fill=obsminor] (obsText) at (7.40,11.15)
    {\textbf{Textual Observation Reduction}};
  \node[subtopic box,fill=obsminor] (obsRegion) at (7.40,9.60)
    {\textbf{Region-Focused Visual Perception}};
  \node[subtopic box,fill=obsminor] (obsHybrid) at (7.40,8.05)
    {\textbf{Observation Enrichment via Parsing}\\\textbf{and Hybridization}};

  \node[subtopic box,fill=memminor] (memSummary) at (7.40,5.95)
    {\textbf{Summary-Based Compression and}\\\textbf{Selective Look-Back}};
  \node[subtopic box,fill=memminor] (memRuntime) at (7.40,4.40)
    {\textbf{Runtime-Level Representation}\\\textbf{Compression}};

  \node[subtopic box,fill=actminor] (actAbstraction) at (7.40,2.30)
    {\textbf{Action Abstraction}};
  \node[subtopic box,fill=actminor] (actVerify) at (7.40,0.75)
    {\textbf{Action Pruning, Verification,}\\\textbf{and Recovery}};
  \node[subtopic box,fill=actminor] (actExplore) at (7.40,-0.80)
    {\textbf{Exploration Control}};

  \node[subtopic box,fill=sysminor] (sysPlanner) at (7.40,-2.90)
    {\textbf{Planner-Side Efficiency}};
  \node[subtopic box,fill=sysminor] (sysRuntime) at (7.40,-4.45)
    {\textbf{System Efficiency Beyond}\\\textbf{the Planner}};

  \node[paper list box,fill=obsminor] (obsTextPapers) at (12.95,11.15)
    {\papercite{Agent-E}{abuelsaad2024agente}; \papercite{Beyond Pixels}{schiepanski2025beyondpixels};\\
    \papercite{LineRetriever}{kerboua2025lineretriever}; \papercite{FocusAgent}{kerboua2025focusagent};\\
    \papercite{Prune4Web}{zhang2025prune4web}; \papercite{Read More, Think More}{enomoto2026readmore}.};
  \node[paper list box,fill=obsminor] (obsRegionPapers) at (12.95,9.60)
    {\papercite{SeeClick}{cheng2024seeclick}; \papercite{Ferret-UI}{you2024ferretui}; \papercite{R-VLM}{park2025rvlm};\\
    \papercite{RegionFocus}{luo2025regionfocus}; \papercite{ShowUI}{lin2025showui}; \papercite{DiMo-GUI}{wu2025dimogui};\\
    \papercite{ScreenSpot-Pro}{li2025screenspotpro}; \papercite{SimpAgent}{chen2025lessismore}.};
  \node[paper list box,fill=obsminor] (obsHybridPapers) at (12.95,8.05)
    {\papercite{Set-of-Mark}{yang2023som}; \papercite{ScreenAI}{baechler2024screenai}; \papercite{OmniParser}{lu2024omniparser};\\
    \papercite{Tree-of-Lens}{fan2024treeoflens}; \papercite{GUI-Actor}{wu2025guiactor}; \papercite{Agent-S}{agashe2024agents};\\
    \papercite{Ferret-UI 2}{li2024ferretui2}; \papercite{UGround}{gou2024uground}; \papercite{Aria-UI}{yang2025aria};\\
    \papercite{Aguvis}{xu2024aguvis}; \papercite{UI-TARS}{qin2025uitars}.};

  \node[paper list box,fill=memminor] (memSummaryPapers) at (12.95,5.95)
    {\papercite{Agent-S}{agashe2024agents}; \papercite{ColorBrowserAgent}{zhou2026colorbrowseragent};\\
    \papercite{GUI-Rise}{liu2025guirise}; \papercite{PAL-UI}{liu2025palui}; \papercite{HiconAgent}{zhou2025hiconagent};\\
    \papercite{SimpAgent}{chen2025lessismore}; \papercite{Read More, Think More}{enomoto2026readmore}.};
  \node[paper list box,fill=memminor] (memRuntimePapers) at (12.95,4.40)
    {\papercite{GUI-KV}{huang2025guikv}; \papercite{ST-Lite}{zhou2026stlite};\\
    \papercite{Continuous Memory}{wu2025continuousmemory}; \papercite{SecAgent}{xie2026secagent}.};

  \node[paper list box,fill=actminor] (actAbstractionPapers) at (12.95,2.30)
    {\papercite{SkillWeaver}{zheng2025skillweaver}; \papercite{PolySkill}{yu2026polyskill};\\
    \papercite{Mobile-Agent-E}{wang2025mobileagente}; \papercite{ActionEngine}{zhong2026actionengine};\\
    \papercite{CoAct-1}{song2025coact1}.};
  \node[paper list box,fill=actminor] (actVerifyPapers) at (12.95,0.75)
    {\papercite{Prune4Web}{zhang2025prune4web}; \papercite{V-Droid}{dai2025vdroid}; \papercite{VeriSafe Agent}{lee2025verisafe};\\
    \papercite{GUI-Shepherd}{chen2025guishepherd}; \papercite{SenseAct}{cai2026senseact};\\
    \papercite{BacktrackAgent}{wu2025backtrackagent}; \papercite{LongHorizonUI}{kang2026longhorizonui}.};
  \node[paper list box,fill=actminor] (actExplorePapers) at (12.95,-0.80)
    {\papercite{LASER}{ma2023laser}; \papercite{Auto-Intent}{kim2024autointent}; \papercite{OpenWebVoyager}{he2024openwebvoyager};\\
    \papercite{GUI-explorer}{xie2025guiexplorer}; \papercite{WebOperator}{dihan2025weboperator};\\
    \papercite{MobileUse}{li2025mobileuse}.};

  \node[paper list box,fill=sysminor] (sysPlannerPapers) at (12.95,-2.90)
    {\papercite{AndroidWorld}{rawles2024androidworld}; \papercite{MMBench-GUI}{wang2025mmbenchgui};\\
    \papercite{OSWorld-Human}{abhyankar2025osworldhuman}; \papercite{UI-R1}{lu2025uir1};\\
    \papercite{Think Twice, Click Once}{tang2025thinktwice}; \papercite{GUI-G1}{zhou2025guig1};\\
    \papercite{AdaGUI-R1}{chen2025adaguir1}; \papercite{MobileUse}{li2025mobileuse}; \papercite{MobileWizard}{lin2025mobilewizard};\\
    \papercite{AgentCPM-GUI}{zhang2025agentcpmgui}; \papercite{Agent S2}{agashe2025agents2};\\
    \papercite{InfiGUIAgent}{liu2026infiguiagent}.};
  \node[paper list box,fill=sysminor] (sysRuntimePapers) at (12.95,-4.45)
    {\papercite{ActionEngine}{zhong2026actionengine}; \papercite{CORE}{fan2025core}; \papercite{GUIGuard}{wang2026guiguard};\\
    \papercite{CoAct-1}{song2025coact1}; \papercite{Agent S2}{agashe2025agents2}; \papercite{IntentCUA}{lee2026intentcua};\\
    \papercite{OS-Symphony}{yang2026ossymphony}; \papercite{LongHorizonUI}{kang2026longhorizonui};\\
    \papercite{UltraCUA}{yang2025ultracua}.};

  \draw[connector] (rootAgent.east) -- (1.65,2.96);
  \draw[connector] (1.65,9.60) -- (1.65,-3.68);

  \draw[connector] (majorObs.west) -- (1.65,9.60);
  \draw[connector] (majorMem.west) -- (1.65,5.18);
  \draw[connector] (majorAct.west) -- (1.65,0.75);
  \draw[connector] (majorSys.west) -- (1.65,-3.68);

  \draw[connector] (4.95,11.15) -- (4.95,8.05);
  \draw[connector] (majorObs.east) -- (4.95,9.60);
  \draw[connector] (4.95,5.95) -- (4.95,4.40);
  \draw[connector] (majorMem.east) -- (4.95,5.18);
  \draw[connector] (4.95,2.30) -- (4.95,-0.80);
  \draw[connector] (majorAct.east) -- (4.95,0.75);
  \draw[connector] (4.95,-2.90) -- (4.95,-4.45);
  \draw[connector] (majorSys.east) -- (4.95,-3.68);

  \draw[connector] (4.95,11.15) -- (obsText.west);
  \draw[connector] (4.95,9.60) -- (obsRegion.west);
  \draw[connector] (4.95,8.05) -- (obsHybrid.west);
  \draw[connector] (4.95,5.95) -- (memSummary.west);
  \draw[connector] (4.95,4.40) -- (memRuntime.west);
  \draw[connector] (4.95,2.30) -- (actAbstraction.west);
  \draw[connector] (4.95,0.75) -- (actVerify.west);
  \draw[connector] (4.95,-0.80) -- (actExplore.west);
  \draw[connector] (4.95,-2.90) -- (sysPlanner.west);
  \draw[connector] (4.95,-4.45) -- (sysRuntime.west);

  \draw[connector] (obsText.east) -- (obsTextPapers.west);
  \draw[connector] (obsRegion.east) -- (obsRegionPapers.west);
  \draw[connector] (obsHybrid.east) -- (obsHybridPapers.west);
  \draw[connector] (memSummary.east) -- (memSummaryPapers.west);
  \draw[connector] (memRuntime.east) -- (memRuntimePapers.west);
  \draw[connector] (actAbstraction.east) -- (actAbstractionPapers.west);
  \draw[connector] (actVerify.east) -- (actVerifyPapers.west);
  \draw[connector] (actExplore.east) -- (actExplorePapers.west);
  \draw[connector] (sysPlanner.east) -- (sysPlannerPapers.west);
  \draw[connector] (sysRuntime.east) -- (sysRuntimePapers.west);
\end{tikzpicture}}
\caption{Survey taxonomy of this review-paper layout rooted at \emph{Efficient GUI Agent}. The left column gives the four major efficiency topics, the middle column preserves the existing subsection taxonomy, and the right column lists the papers integrated under each subtopic.}
\label{fig:taxonomy_overview}
\end{figure*}
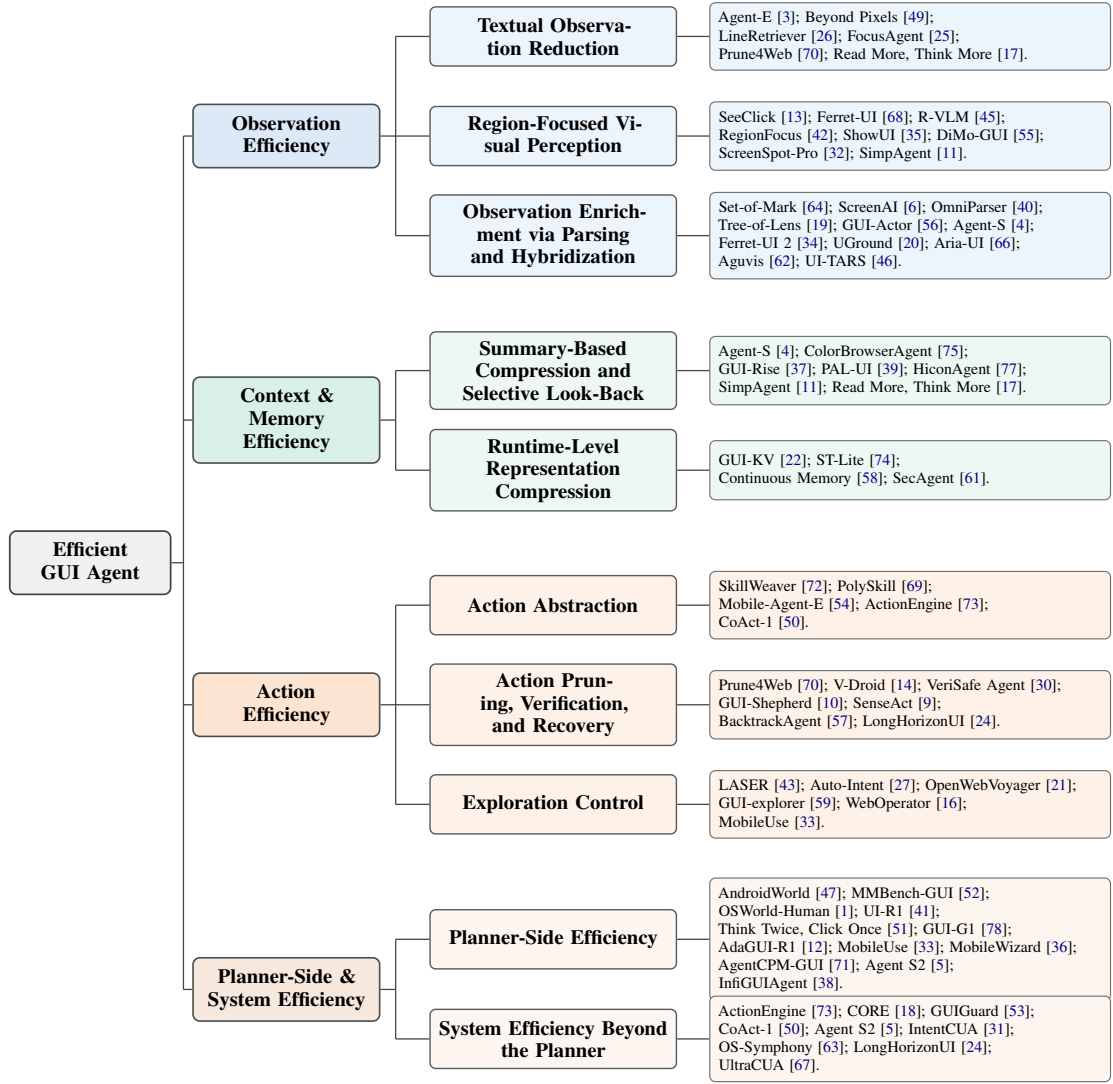

\section{Observation Efficiency}
\label{sec:observation_efficiency}

Observation efficiency concerns how a GUI agent acquires a decision-sufficient representation of the current interface while minimizing redundancy, grounding ambiguity, and downstream reasoning cost. In GUI environments, inefficiency may arise from overly long textual interface representations, visually cluttered screenshots, or weakly structured multimodal observations. 
\subsection{Textual Observation Reduction}

For agents operating on the Document Object Model (DOM) or the accessibility tree (AxTree), the primary inefficiency arises from an overabundance of textual interface structure. Real web pages often expose large interface trees containing decorative nodes, repeated text spans, layout artifacts, and task-irrelevant branches, which substantially increase computational cost without improving action prediction~\cite{abuelsaad2024agente}. Agent-E made this issue explicit in practical web-agent design by advocating flexible DOM distillation and denoising as part of a hierarchical browser-agent architecture, where raw DOM inputs can reach up to 800k tokens and typical tasks require 150--220 seconds and about 25 LLM calls~\cite{abuelsaad2024agente}. Beyond Pixels studies the issue directly through DOM downsampling, showing that aggressively compressed DOM snapshots can remain around the \(10^3\)-token order while preserving useful hierarchical signals for downstream decision making~\cite{schiepanski2025beyondpixels}. A more selective direction formulates reduction as retrieval rather than uniform truncation. LineRetriever argues that the most useful lines are those that support future navigation decisions rather than those that are merely semantically similar to the goal text, reporting observation reductions of 61\%, 72\%, and 73\% while retrieving up to 10 chunks of 100 tokens~\cite{kerboua2025lineretriever}. FocusAgent selectively retrieves task-relevant AxTree lines at each step, achieving more than 50\% average AxTree reduction, often more than 80\%, while capping context at 2k tokens~\cite{kerboua2025focusagent}. Prune4Web pushes this line further by moving pruning from inference-time reading into explicit executable filtering programs, with examples where DOM trees of 10k--100k tokens and more than 500 candidate elements are reduced to fewer than 20 actionable candidates, thereby reducing both observation length and grounding search complexity~\cite{zhang2025prune4web}.

\subsection{Region-Focused Visual Perception}
When the primary observation is a screenshot, inefficiency is mainly an allocation problem: GUI images contain small actionable targets, repeated widgets, and irrelevant background regions. SeeClick established the feasibility of screenshot-only GUI agents while exposing visual grounding as a major bottleneck, although it does not report directly comparable token, latency, memory, or step savings~\cite{cheng2024seeclick}. Ferret-UI and R-VLM address this bottleneck by reallocating visual resolution through sub-image partitioning, zoomed region proposals, and region-aware grounding objectives; Ferret-UI splits each screen into two sub-images for any-resolution processing, while R-VLM reports about 5.6 seconds per sample and up to \(2\times\) inference-latency cost, showing that region proposals may shift cost into extra inference~\cite{you2024ferretui,park2025rvlm}. RegionFocus and DiMo-GUI make this focusing process adaptive at inference time, progressively zooming into task-relevant or ambiguous regions instead of encoding the full screen uniformly; this improves visual allocation but can introduce new overhead, with RegionFocus reporting 66.8\% average trajectory overhead and a 19.74\% step-count increase, and DiMo-GUI using up to seven zoom iterations~\cite{luo2025regionfocus,wu2025dimogui}. ShowUI applies the same principle at the token level, selecting UI-relevant visual tokens to remove 33\% redundant visual tokens and obtain a \(1.4\times\) training speedup~\cite{lin2025showui}. ScreenSpot-Pro and SimpAgent further show why such selective perception matters: professional high-resolution GUI screens often exceed \(3k\times2k\) resolution, while SimpAgent compresses history images into 64 tokens, uses 10--20-token action outputs, and reports a 27\% FLOP reduction in the LLM branch~\cite{li2025screenspotpro,chen2025lessismore}. 
\subsection{Observation Enrichment via Parsing and Hybridization}
Observation efficiency is not only input reduction; it also depends on whether the observation is immediately actionable. Screenshots preserve rendered context but lack structure, while DOM or AxTree representations are compact yet often incomplete, noisy, or misaligned with the rendered interface. Set-of-Mark and ScreenAI address this gap by adding lightweight referential or textual annotations to screen elements; ScreenAI also illustrates the scale of screen-understanding backbones, reporting 670M, 2B, and 5B model variants and input resolution up to \(812^2\)~\cite{yang2023som,baechler2024screenai}. OmniParser and Tree-of-Lens go further by converting screenshots or pointed regions into semi-structured region, function, and layout representations that expose content and spatial relations for downstream reasoning; OmniParser adds two parser models and can restrict downstream reasoning to the top-50 relevant elements~\cite{lu2024omniparser,fan2024treeoflens}. GUI-Actor complements this parser-style enrichment with coordinate-free action regions and a grounding verifier, adding about 20M parameters for a 2B model and 100M parameters for a 7B model so that action candidates can be generated in one forward pass while reducing brittle coordinate decoding~\cite{wu2025guiactor}. Hybrid systems such as Agent-S and Ferret-UI 2 combine rendered visual context with accessibility or cross-platform structural signals, whereas UGround, Aria-UI, Aguvis, and UI-TARS show that strong screenshot-first pipelines can also support grounding, planning, and action prediction without relying on full DOM or AxTree input: UGround uses about two-thirds of the visual tokens required by a fixed \(1344\times1344\) setting,  Aguvis reduces text-agent inputs from roughly 4k--6k tokens per step to 1{,}196 tokens for a 70\% input-token reduction ~\cite{agashe2024agents,li2024ferretui2,gou2024uground,yang2025aria,xu2024aguvis,qin2025uitars}. 
\section{Context and Memory Efficiency}
\label{sec:memory_efficiency}

Context and memory efficiency concerns how agents preserve, access, and update historical information over long interaction horizons without allowing context growth to dominate runtime and memory consumption. 

\subsection{Summary-Based Compression and Selective Look-Back}
As GUI trajectories grow longer, replaying past screenshots, actions, and reasoning traces becomes both costly and noisy. Agent-S, ColorBrowserAgent, and GUI-Rise address this problem by replacing raw trajectory replay with compact task-progress memories, progressive summaries, or progress-aware summaries trained to support later action prediction; ColorBrowserAgent, for example, caps the interaction horizon at 30 steps, highlighting the need to prevent history growth from becoming unbounded~\cite{agashe2024agents,zhou2026colorbrowseragent,liu2025guirise}. More recent systems make this compression selective rather than purely lossy. PAL-UI combines dual-level summarization with active look-back, while HiconAgent and SimpAgent reduce redundant history through dynamic context sampling, anchor-guided compression, or consistency-guided pruning; HiconAgent reports 25.21T FLOPs for compressed history compared with 35.75T for an uncompressed 3B setting and 62.31T for a 7B setting~\cite{liu2025palui,zhou2025hiconagent,chen2025lessismore}. Read More, Think More~\cite{enomoto2026readmore} showing that diff-based history can be more token-efficient than replaying full prior observations, especially when WorkArena HTML pages can reach 40k--500k tokens under 4-step or 9-step look-back settings.

\subsection{Runtime-Level Representation Compression}

A second family of methods addresses the memory bottleneck directly at the representation and serving level. These methods optimize \emph{how} the preserved information is represented and reused during inference. GUI-KV exploits the spatial and temporal redundancy of GUI trajectories, tailoring compression policies to GUI-specific attention patterns; it reports 38.9\% fewer MFLOPs per decoded token at five screenshots, uses 5--20\% cache budgets, and notes that five screenshots can exceed 80GB of GPU memory~\cite{huang2025guikv}. ST-Lite similarly targets the cache bottleneck with a training-free KV-compression strategy designed for long-horizon GUI workloads, operating at 10--20\% cache budgets and reporting a \(2.45\times\) decoding speedup and \(1.40\times\) end-to-end speedup, although prefill remains close to \(1.0\times\), indicating that the savings are concentrated mainly on the decode side~\cite{zhou2026stlite}. These methods are especially important because they improve scalability even when the high-level memory policy remains unchanged.

Other work reduces context cost by replacing symbolic histories with denser internal representations. Auto-scaling Continuous Memory for GUI Agent replaces long textual summaries with fixed-length continuous memory representations that preserve fine-grained visual information, compressing each trajectory into eight embeddings even when raw trajectories exceed 15k tokens, with a reported data-collection cost of about \$4k and tuning of 1.2\% of parameters~\cite{wu2025continuousmemory}. SecAgent takes a lighter-weight route by distilling prior screenshots and actions into concise semantic context for efficient mobile control~\cite{xie2026secagent}.
\section{Action Efficiency}
\label{sec:action_efficiency}

Even when model inference is efficient, GUI agents can remain impractical because they waste interaction steps. Action efficiency studies how to complete tasks with fewer actions, fewer irreversible mistakes, and less unproductive exploration. 

\subsection{Action Abstraction}
Action abstraction reduces step count by lifting repeated primitive operations into reusable skills, routines, or programs. In the web setting, SkillWeaver discovers reusable website procedures and distills them into lightweight callable APIs~\cite{zheng2025skillweaver}. PolySkill sharpens that idea from a transfer perspective by separating a skill's abstract goal from its site-specific implementation, with learned functions typically covering 2--5 GUI steps and allowing web agents to reuse skills across seen and unseen websites with fewer redundant exploratory actions~\cite{yu2026polyskill}. Mobile-Agent-E adopts a reusable \emph{Shortcuts} with explicit preconditions, turning recurrent subtasks into more efficient and more robust routines~\cite{wang2025mobileagente}. These methods share a stable intuition: if a control pattern recurs, the agent should not keep paying the full planning cost every time it appears.

At a more programmatic extreme, ActionEngine replaces repeated reactive planning with state-machine memory and executable programs, showing that frequent GUI interaction patterns can be compiled into reusable control structures; this compilation reduces cost from \$0.71 to \$0.06, latency from 237.5 to 118.3 seconds, input tokens from 62.3k to 8.1k, and model calls from 10.2 to 1.8~\cite{zhong2026actionengine}. CoAct-1 expands abstraction beyond pure GUI actions by allowing coding to function as an execution modality, letting the system dynamically delegate a subtask either to a GUI operator or to a programmer agent that writes and executes code; this routing reduces average steps to 10.15, compared with 15.22 for GTA-1 and 14.90 for UI-TARS~\cite{song2025coact1}. This matters for action efficiency because some subtasks, especially OS-level file manipulation or data processing, are inefficient when expressed as long GUI-only trajectories.
\subsection{Action Pruning, Verification, and Recovery}

A second route to action efficiency is to reduce or validate candidate actions before expensive reasoning or irreversible execution. Prune4Web reduces both observation size and downstream action search by pruning DOM candidates early, with examples where more than 500 DOM elements shrink to fewer than 20 candidates~\cite{zhang2025prune4web}. V-Droid, ALTP and VeriSafe Agent score or verify candidate actions before execution to improve deployment efficiency and intent alignment: V-Droid reports 2.6k--8.9k input tokens, 0.7 seconds per decision, and 4.3 seconds per step compared with typical mobile agents above 20 seconds per step, while VeriSafe Agent shows recovery examples requiring 1--3 actions and notes that tasks above 10 steps can exceed \$1~\cite{dai2025vdroid,lee2025verisafe,bai2025localinformationmattersinference}. GUI-Shepherd learns process-level rewards that can later be reused as a verifier signal during inference, although verifier-call and latency metrics are not yet reported separately~\cite{chen2025guishepherd}. SenseAct structures actions through typed commitments and post-condition checks, reducing UI exposure by 65.51\% and reducing the need to repeatedly consult a large VLM for every low-level execution decision~\cite{cai2026senseact}.

Once actions are executed in long-horizon settings, recovery becomes equally important. BacktrackAgent explicitly introduces backtracking, together with verifier, judger, and reflector modules, to mitigate cascading failures after early mistakes~\cite{wu2025backtrackagent}. LongHorizonUI combines reflective decision making with rollback-oriented execution over average trajectories of 24.6 steps and maximum trajectories of 37 steps~\cite{kang2026longhorizonui}. The shared lesson is that action efficiency is not only about choosing fewer actions. It is also about failing in ways that are detectable, reversible, and cheap to correct. 

\subsection{Exploration Control}

A third source of inefficiency is unstructured exploration in unfamiliar interfaces. LASER models web interaction as state-space exploration with explicit backtracking, making recovery from off-trajectory decisions more systematic than in forward-only prompting schemes \cite{ma2023laser}. Auto-Intent distills compact intents from demonstrations and uses them to guide self-exploration, so that search is constrained by higher-level latent goals rather than by unrestricted trial and error \cite{kim2024autointent}. OpenWebVoyager propose an exploration--feedback--optimization loop, treating exploration as a continual self-improvement process rather than as a one-off phase \cite{he2024openwebvoyager}. GUI-explorer further enriches this line by autonomously mining transition-aware knowledge and function-aware trajectories \cite{xie2025guiexplorer}. WebOperator  combines pre-execution filtering, best-first search, and safe backtracking so that the agent does not pay the full price of blind rollout in complex web tasks \cite{dihan2025weboperator}.MobileUse propose  a system where proactive exploration is invoked to handle cold-start mobile environments rather than as an always-on behavior \cite{li2025mobileuse}. Across these works, the shared lesson is that exploration becomes action-efficient only when it is guided by reusable structure, predicted intent, or learned transition logic.

\section{Planner-Side and System Efficiency}
\label{sec:planner_system_efficiency}

Planner-side efficiency concerns the computational burden inside the decision loop, while system efficiency concerns the broader runtime around the planner, including memory reuse, privacy-aware routing, and orchestration across heterogeneous components.

\subsection{Planner-Side Efficiency}

Recent benchmark work makes clear that efficiency is not reducible to final task success. MMBench-GUI introduce an explicitly efficiency-aware evaluation perspective across multiple platforms, using a 50-step budget and reporting redundant-step cost values of 7--8, privacy noise of 40\%, and specialist cost of 16\%~\cite{wang2025mmbenchgui}. OSWorld-Human sharpens the diagnosis further by showing that planning and reflection dominate end-to-end latency in computer-use agents, with agents requiring \(2.7\times\)--\(4.3\times\) more steps than human-derived trajectories and reflection accounting for 76\%--96\% of task latency~\cite{abhyankar2025osworldhuman}. These benchmark-driven analyses are important because they motivate planner-side efficiency as a first-class systems objective rather than an afterthought.

Methodologically, UI-R1 shows that rule-based reinforcement learning can improve GUI action prediction efficiently even in relatively compact models, using only 136 training samples and about 8 hours on 8 RTX 4090 GPUs, while also exposing OOM risk under high-pixel settings~\cite{lu2025uir1}. Think Twice, Click Once argues that grounding and decision making should not use uniform reasoning depth; instead, fast and slow modes should be deployed selectively according to task difficulty, even though invoking slow thinking raises processing time from 2.6 to 5.4 seconds and its data include 300k examples with 150k slow samples~\cite{tang2025thinktwice}. GUI-G1 analyzes R1-style GUI grounding pipelines and finds that longer chains of thought can hurt rather than help. AdaGUI-R1 reports a similar lesson from adaptive reasoning scheduling, reducing unnecessary reasoning tokens by 40\% while adding 23.5\% FLOPs under its harder-example schedule, suggesting that selective thinking is likely to remain central in mobile and cross-app control~\cite{chen2025adaguir1}.

Other methods factorize planning rather than merely shortening it. MobileUse uses hierarchical reflection invoked on demand rather than at every step, reducing reflection overhead to 10\% after exploration that can cost up to 100 steps per app at 19.5 seconds per step~\cite{li2025mobileuse}. MobileWizard combines structured reasoning with progressive reinforcement learning to achieve strong performance from modest data budgets, using 24.5k public trajectories plus 300 remedial trajectories and fewer than 50k trajectories overall~\cite{lin2025mobilewizard}. AgentCPM-GUI emphasizes efficient deployment through grounding-aware training and a compact action space, trained on 470k atomic steps with about 8.5 steps per trajectory, while retaining the last four actions/images and using a max\_new\_tokens setting of 2048~\cite{zhang2025agentcpmgui}. Agent S2 reframes planner efficiency compositionally, delegating different cognitive roles across generalist and specialist modules to avoid overloading a single monolithic planner with every subproblem~\cite{agashe2025agents2}. InfiGUIAgent adds another perspective by showing that a compact generalist model can integrate native reasoning and reflection without fully externalizing every planner role into a separate subsystem~\cite{liu2026infiguiagent}. 

\subsection{System Efficiency Beyond the Planner}

System efficiency depends on the runtime around the planner, not only on the planner itself. ActionEngine amortizes repeated online reasoning through state-machine memory and program synthesis, while CoAct-1 routes subtasks between GUI interaction and code execution when different backends offer lower cost \cite{zhong2026actionengine,song2025coact1}. CORE and GUIGuard make deployment topology part of the efficiency problem by splitting work across local/cloud models and adding explicit privacy-recognition or privacy-protection stages \cite{fan2025core,wang2026guiguard}. IntentCUA reduces redundant replanning through intent-level abstractions and shared plan memory across collaborative desktop workflows \cite{lee2026intentcua}. OS-Symphony and LongHorizonUI extend this systems view to long-horizon robustness, combining reflection, memory, tutorial retrieval, element-indexed perception, or rollback-based execution to keep trajectories recoverable in unseen or sustained tasks \cite{yang2026ossymphony,kang2026longhorizonui}. The central lesson is that efficient GUI agency requires joint accounting of observation design, memory placement, execution routing, verification, and orchestration cost across the full runtime.
\section{Open Challenges and Future Directions}
\label{sec:open_challenges}
Several open problems follow from this systems view. First, the field still lacks honest efficiency accounting: many papers report local savings in tokens, actions, or module latency, but benchmarks often stop at success rate or aggregate completion, making cross-paper comparison difficult. OSWorld-Human and MMBench-GUI move in the right direction by profiling efficiency more directly, yet verifier calls, parser overhead, and multi-agent orchestration still lack a shared pricing framework \cite{abhyankar2025osworldhuman,wang2025mmbenchgui}. 
The second challenge is that there are fewer GUI-agent benchmarks. Future benchmarks should report peak GPU memory, prefill and decode latency, MFLOPs per decoded token, GPU-hours for training or search, and success-normalized GPU cost, so that the community can evaluate whether an agent is deployable under realistic serving constraints rather than merely efficient in tokens or steps.

\clearpage
\bibliography{example_paper}

@article{abuelsaad2024agente,
  title={Agent-E: From Autonomous Web Navigation to Foundational Design Principles in Agentic Systems},
  author={Abuelsaad, Tamer and Akkil, Deepak and Dey, Prasenjit and Jagmohan, Ashish and Vempaty, Aditya and Kokku, Ravi},
  journal={arXiv preprint arXiv:2407.13032},
  year={2024},
  url={https://arxiv.org/abs/2407.13032}
}

@misc{bai2025localinformationmattersinference,
      title={Local Information Matters: Inference Acceleration For Grounded Conversation Generation Models Through Adaptive Local-Aware Token Pruning}, 
      author={Bizhe Bai and Jianjian Cao and Yadan Luo and Tao Chen},
      year={2025},
      eprint={2503.23959},
      archivePrefix={arXiv},
      primaryClass={cs.CV},
      url={https://arxiv.org/abs/2503.23959}, 
}

@article{schiepanski2025beyondpixels,
  title={Beyond Pixels: Exploring DOM Downsampling for LLM-Based Web Agents},
  author={Schiepanski, Thassilo M. and Piël, Nicholas},
  journal={arXiv preprint arXiv:2508.04412},
  year={2025},
  url={https://arxiv.org/abs/2508.04412}
}

@misc{abhyankar2026osworldhumanbenchmarkingefficiencycomputeruse,
      title={OSWorld-Human: Benchmarking the Efficiency of Computer-Use Agents}, 
      author={Reyna Abhyankar and Qi Qi and Yiying Zhang},
      year={2026},
      eprint={2506.16042},
      archivePrefix={arXiv},
      primaryClass={cs.AI},
      url={https://arxiv.org/abs/2506.16042}, 
}

@article{kerboua2025lineretriever,
  title={LineRetriever: Planning-Aware Observation Reduction for Web Agents},
  author={Kerboua, Imene and Omidi Shayegan, Sahar and Thakkar, Megh and Lù, Xing Han and Caccia, Massimo and Eglin, Véronique and Aussem, Alexandre and Espinas, Jérémy and Lacoste, Alexandre},
  journal={arXiv preprint arXiv:2507.00210},
  year={2025},
  url={https://arxiv.org/abs/2507.00210}
}

@article{kerboua2025focusagent,
  title={FocusAgent: Simple Yet Effective Ways of Trimming the Large Context of Web Agents},
  author={Kerboua, Imene and Omidi Shayegan, Sahar and Lù, Xing Han and Boisvert, Léo and Thakkar, Megh and Caccia, Massimo and Espinas, Jérémy and Aussem, Alexandre and Eglin, Véronique and Lacoste, Alexandre},
  journal={arXiv preprint arXiv:2510.03204},
  year={2025},
  url={https://arxiv.org/abs/2510.03204}
}

@article{zhang2025prune4web,
  title={Prune4Web: DOM Tree Pruning Programming for Web Agent},
  author={Zhang, Jiayuan and Chen, Kaiquan and Lu, Zhihao and Zhou, Enshen and Yu, Qian and Zhang, Jing},
  journal={arXiv preprint arXiv:2511.21398},
  year={2025},
  url={https://arxiv.org/abs/2511.21398}
}

@inproceedings{cheng2024seeclick,
  title={SeeClick: Harnessing GUI Grounding for Advanced Visual GUI Agents},
  author={Cheng, Kanzhi and Sun, Qiushi and Chu, Yougang and Xu, Fangzhi and Li, YanTao and Zhang, Jianbing and Wu, Zhiyong},
  booktitle={Proceedings of the 62nd Annual Meeting of the Association for Computational Linguistics (Volume 1: Long Papers)},
  year={2024},
  pages={9313--9332},
  url={https://aclanthology.org/2024.acl-long.505/}
}

@article{you2024ferretui,
  title={Ferret-UI: Grounded Mobile UI Understanding with Multimodal LLMs},
  author={You, Keen and Zhang, Haotian and Schoop, Eldon and Weers, Floris and Swearngin, Amanda and Nichols, Jeffrey and Yang, Yinfei and Gan, Zhe},
  journal={arXiv preprint arXiv:2404.05719},
  year={2024},
  url={https://arxiv.org/abs/2404.05719}
}

@article{fan2024treeoflens,
  title={Read Anywhere Pointed: Layout-aware GUI Screen Reading with Tree-of-Lens Grounding},
  author={Fan, Yue and Ding, Lei and Kuo, Ching-Chen and Jiang, Shan and Zhao, Yang and Guan, Xinze and Yang, Jie and Zhang, Yi and Wang, Xin Eric},
  journal={arXiv preprint arXiv:2406.19263},
  year={2024},
  url={https://arxiv.org/abs/2406.19263}
}

@article{luo2025regionfocus,
  title={Visual Test-time Scaling for GUI Agent Grounding},
  author={Luo, Tiange and Logeswaran, Lajanugen and Johnson, Justin and Lee, Honglak},
  journal={arXiv preprint arXiv:2505.00684},
  year={2025},
  url={https://arxiv.org/abs/2505.00684}
}

@article{yang2023som,
  title={Set-of-Mark Prompting Unleashes Extraordinary Visual Grounding in GPT-4V},
  author={Yang, Jianwei and Zhang, Hao and Li, Feng and Zou, Xueyan and Li, Chunyuan and Gao, Jianfeng},
  journal={arXiv preprint arXiv:2310.11441},
  year={2023},
  url={https://arxiv.org/abs/2310.11441}
}

@article{baechler2024screenai,
  title={ScreenAI: A Vision-Language Model for UI and Infographics Understanding},
  author={Baechler, Gilles and Sunkara, Srinivas and Wang, Maria and Zubach, Fedir and Mansoor, Hassan and Etter, Vincent and Cărbune, Victor and Lin, Jason and Chen, Jindong and Sharma, Abhanshu},
  journal={arXiv preprint arXiv:2402.04615},
  year={2024},
  url={https://arxiv.org/abs/2402.04615}
}

@article{lu2024omniparser,
  title={OmniParser for Pure Vision Based GUI Agent},
  author={Lu, Yadong and Yang, Jianwei and Shen, Yelong and Awadallah, Ahmed},
  journal={arXiv preprint arXiv:2408.00203},
  year={2024},
  url={https://arxiv.org/abs/2408.00203}
}

@article{agashe2024agents,
  title={Agent S: An Open Agentic Framework that Uses Computers Like a Human},
  author={Agashe, Saaket and Han, Jiuzhou and Gan, Shuyu and Yang, Jiachen and Li, Ang and Wang, Xin Eric},
  journal={arXiv preprint arXiv:2410.08164},
  year={2024},
  url={https://arxiv.org/abs/2410.08164}
}

@article{li2024ferretui2,
  title={Ferret-UI 2: Mastering Universal User Interface Understanding Across Platforms},
  author={Li, Zhangheng and You, Keen and Zhang, Haotian and Feng, Di and Agrawal, Harsh and Li, Xiujun and Moorthy, Mohana Prasad Sathya and Nichols, Jeff and Yang, Yinfei and Gan, Zhe},
  journal={arXiv preprint arXiv:2410.18967},
  year={2024},
  url={https://arxiv.org/abs/2410.18967}
}

@article{gou2024uground,
  title={Navigating the Digital World as Humans Do: Universal Visual Grounding for GUI Agents},
  author={Gou, Boyu and Wang, Ruohan and Zheng, Boyuan and Xie, Yanan and Chang, Cheng and Shu, Yiheng and Sun, Huan and Su, Yu},
  journal={arXiv preprint arXiv:2410.05243},
  year={2024},
  url={https://arxiv.org/abs/2410.05243}
}

@article{xu2024aguvis,
  title={Aguvis: Unified Pure Vision Agents for Autonomous GUI Interaction},
  author={Xu, Yiheng and Wang, Zekun and Wang, Junli and Lu, Dunjie and Xie, Tianbao and Saha, Amrita and Sahoo, Doyen and Yu, Tao and Xiong, Caiming},
  journal={arXiv preprint arXiv:2412.04454},
  year={2024},
  url={https://arxiv.org/abs/2412.04454}
}

@inproceedings{yang2025aria,
  title={Aria-UI: Visual Grounding for GUI Instructions},
  author={Yang, Yuhao and Wang, Yue and Li, Dongxu and Luo, Ziyang and Chen, Bei and Huang, Chao and Li, Junnan},
  booktitle={Findings of the Association for Computational Linguistics: ACL 2025},
  year={2025},
  pages={22418--22433},
  doi={10.18653/v1/2025.findings-acl.1152},
  url={https://aclanthology.org/2025.findings-acl.1152/}
}

@article{qin2025uitars,
  title={UI-TARS: Pioneering Automated GUI Interaction with Native Agents},
  author={Qin, Yujia and Ye, Yining and Fang, Junjie and Wang, Haoming and Liang, Shihao and Tian, Shizuo and Zhang, Junda and Li, Jiahao and Li, Yunxin and Huang, Shijue and Zhong, Wanjun and Li, Kuanye and Yang, Jiale and Miao, Yu and Lin, Woyu and Liu, Longxiang and Jiang, Xu and Ma, Qianli and Li, Jingyu and Xiao, Xiaojun and Cai, Kai and Li, Chuang and Zheng, Yaowei and Jin, Chaolin and Li, Chen and Zhou, Xiao and Wang, Minchao and Chen, Haoli and Li, Zhaojian and Yang, Haihua and Liu, Haifeng and Lin, Feng and Peng, Tao and Liu, Xin and Shi, Guang},
  journal={arXiv preprint arXiv:2501.12326},
  year={2025},
  url={https://arxiv.org/abs/2501.12326}
}

@article{liu2025palui,
  title={PAL-UI: Planning with Active Look-back for Vision-Based GUI Agents},
  author={Liu, Zikang and Li, Junyi and Zhao, Wayne Xin and Gao, Dawei and Li, Yaliang and Wen, Ji-rong},
  journal={arXiv preprint arXiv:2510.00413},
  year={2025},
  url={https://arxiv.org/abs/2510.00413}
}

@article{zhou2025hiconagent,
  title={HiconAgent: History Context-aware Policy Optimization for GUI Agents},
  author={Zhou, Xurui and Chen, Gongwei and Xie, Yuquan and Li, Zaijing and Zhou, Kaiwen and Wang, Shuai and Yang, Shuo and Tian, Zhuotao and Shao, Rui},
  journal={arXiv preprint arXiv:2512.01763},
  year={2025},
  url={https://arxiv.org/abs/2512.01763}
}

@article{huang2025guikv,
  title={GUI-KV: Efficient GUI Agents via KV Cache with Spatio-Temporal Awareness},
  author={Huang, Kung-Hsiang and Qiu, Haoyi and Dai, Yutong and Xiong, Caiming and Wu, Chien-Sheng},
  journal={arXiv preprint arXiv:2510.00536},
  year={2025},
  url={https://arxiv.org/abs/2510.00536}
}

@article{wu2025continuousmemory,
  title={Auto-scaling Continuous Memory for GUI Agent},
  author={Wu, Wenyi and Zhou, Kun and Yuan, Ruoxin and Yu, Vivian and Wang, Stephen and Hu, Zhiting and Huang, Biwei},
  journal={arXiv preprint arXiv:2510.09038},
  year={2025},
  url={https://arxiv.org/abs/2510.09038}
}

@article{zhou2026stlite,
  title={Efficient Long-Horizon GUI Agents via Training-Free KV Cache Compression},
  author={Zhou, Bowen and Xu, Zhou and Li, Wanli and Xiao, Jingyu and Wang, Haoqian},
  journal={arXiv preprint arXiv:2603.00188},
  year={2026},
  url={https://arxiv.org/abs/2603.00188}
}

@article{xie2026secagent,
  title={SecAgent: Efficient Mobile GUI Agent with Semantic Context},
  author={Xie, Yiping and Chen, Song and Xing, Jingxuan and Jiang, Wei and Zhu, Zekun and Wang, Yingyao and Bu, Pi and Song, Jun and Jiang, Yuning and Zheng, Bo},
  journal={arXiv preprint arXiv:2603.08533},
  year={2026},
  url={https://arxiv.org/abs/2603.08533}
}

@article{zheng2025skillweaver,
  title={SkillWeaver: Web Agents can Self-Improve by Discovering and Honing Skills},
  author={Zheng, Boyuan and Fatemi, Michael Y. and Jin, Xiaolong and Wang, Zora Zhiruo and Gandhi, Apurva and Song, Yueqi and Gu, Yu and Srinivasa, Jayanth and Liu, Gaowen and Neubig, Graham and Su, Yu},
  journal={arXiv preprint arXiv:2504.07079},
  year={2025},
  url={https://arxiv.org/abs/2504.07079}
}

@article{wang2025mobileagente,
  title={Mobile-Agent-E: Self-Evolving Mobile Assistant for Complex Tasks},
  author={Wang, Zhenhailong and Xu, Haiyang and Wang, Junyang and Zhang, Xi and Yan, Ming and Zhang, Ji and Huang, Fei and Ji, Heng},
  journal={arXiv preprint arXiv:2501.11733},
  year={2025},
  url={https://arxiv.org/abs/2501.11733}
}

@article{dai2025vdroid,
  title={Advancing Mobile GUI Agents: A Verifier-Driven Approach to Practical Deployment},
  author={Dai, Gaole and Jiang, Shiqi and Cao, Ting and Li, Yuanchun and Yang, Yuqing and Tan, Rui and Li, Mo and Qiu, Lili},
  journal={arXiv preprint arXiv:2503.15937},
  year={2025},
  url={https://arxiv.org/abs/2503.15937}
}

@inproceedings{wu2025backtrackagent,
  title={BacktrackAgent: Enhancing GUI Agent with Error Detection and Backtracking Mechanism},
  author={Wu, Qinzhuo and Gao, Pengzhi and Liu, Wei and Luan, Jian},
  booktitle={Proceedings of the 2025 Conference on Empirical Methods in Natural Language Processing},
  year={2025},
  pages={4250--4272},
  doi={10.18653/v1/2025.emnlp-main.212},
  url={https://aclanthology.org/2025.emnlp-main.212/}
}

@article{ma2023laser,
  title={LASER: LLM Agent with State-Space Exploration for Web Navigation},
  author={Ma, Kaixin and Zhang, Hongming and Wang, Hongwei and Pan, Xiaoman and Yu, Dong},
  journal={arXiv preprint arXiv:2309.08172},
  year={2023},
  url={https://arxiv.org/abs/2309.08172}
}

@article{kim2024autointent,
  title={Auto-Intent: Automated Intent Discovery and Self-Exploration for Large Language Model Web Agents},
  author={Kim, Jaekyeom and Kim, Dong-Ki and Logeswaran, Lajanugen and Sohn, Sungryull and Lee, Honglak},
  journal={arXiv preprint arXiv:2410.22552},
  year={2024},
  url={https://arxiv.org/abs/2410.22552}
}

@article{he2024openwebvoyager,
  title={OpenWebVoyager: Building Multimodal Web Agents via Iterative Real-World Exploration, Feedback and Optimization},
  author={He, Hongliang and Yao, Wenlin and Ma, Kaixin and Yu, Wenhao and Zhang, Hongming and Fang, Tianqing and Lan, Zhenzhong and Yu, Dong},
  journal={arXiv preprint arXiv:2410.19609},
  year={2024},
  url={https://arxiv.org/abs/2410.19609}
}

@article{rawles2024androidworld,
  title={AndroidWorld: A Dynamic Benchmarking Environment for Autonomous Agents},
  author={Rawles, Christopher and Clinckemaillie, Sarah and Chang, Yifan and Waltz, Jonathan and Lau, Gabrielle and Fair, Marybeth and Li, Alice and Bishop, William and Li, Wei and Campbell-Ajala, Folawiyo and Toyama, Daniel and Berry, Robert and Tyamagundlu, Divya and Lillicrap, Timothy and Riva, Oriana},
  journal={arXiv preprint arXiv:2405.14573},
  year={2024},
  url={https://arxiv.org/abs/2405.14573}
}

@article{lu2025uir1,
  title={UI-R1: Enhancing Efficient Action Prediction of GUI Agents by Reinforcement Learning},
  author={Lu, Zhengxi and Chai, Yuxiang and Guo, Yaxuan and Yin, Xi and Liu, Liang and Wang, Hao and Xiao, Han and Ren, Shuai and Xiong, Guanjing and Li, Hongsheng},
  journal={arXiv preprint arXiv:2503.21620},
  year={2025},
  url={https://arxiv.org/abs/2503.21620}
}

@article{tang2025thinktwice,
  title={Think Twice, Click Once: Enhancing GUI Grounding via Fast and Slow Systems},
  author={Tang, Fei and Shen, Yongliang and Zhang, Hang and Chen, Siqi and Hou, Guiyang and Zhang, Wenqi and Zhang, Wenqiao and Song, Kaitao and Lu, Weiming and Zhuang, Yueting},
  journal={arXiv preprint arXiv:2503.06470},
  year={2025},
  url={https://arxiv.org/abs/2503.06470}
}

@misc{li2025mobileuse,
  title={MobileUse: A Hierarchical Reflection-Driven GUI Agent for Autonomous Mobile Operation},
  author={Li, Ning and Qu, Xiangmou and Zhou, Jiamu and Wang, Jun and Wen, Muning and Du, Kounianhua and Lou, Xingyu and Peng, Qiuying and Wang, Jun and Zhang, Weinan},
  year={2025},
  howpublished={OpenReview, NeurIPS 2025 poster},
  url={https://openreview.net/forum?id=KR6tnkb6h4}
}

@article{zhong2026actionengine,
  title={ActionEngine: From Reactive to Programmatic GUI Agents via State Machine Memory},
  author={Zhong, Hongbin and Faisal, Fazle and Fran{\c{c}}a, Luis and Leesatapornwongsa, Tanakorn and Szekeres, Adriana and Rong, Kexin and Nath, Suman},
  journal={arXiv preprint arXiv:2602.20502},
  year={2026},
  url={https://arxiv.org/abs/2602.20502}
}

@inproceedings{fan2025core,
  title={CORE: Reducing UI Exposure in Mobile Agents via Collaboration Between Cloud and Local LLMs},
  author={Fan, Gucongcong and Niu, Chaoyue and Lyu, Chengfei and Wu, Fan and Chen, Guihai},
  booktitle={Advances in Neural Information Processing Systems},
  year={2025},
  url={https://openreview.net/forum?id=klOr9y9nMU}
}

@article{wang2026guiguard,
  title={GUIGuard: Toward a General Framework for Privacy-Preserving GUI Agents},
  author={Wang, Yanxi and Zhang, Zhiling and Zhou, Wenbo and Zhang, Weiming and Zhang, Jie and Zhu, Qiannan and Shi, Yu and Zheng, Shuxin and He, Jiyan},
  journal={arXiv preprint arXiv:2601.18842},
  year={2026},
  url={https://arxiv.org/abs/2601.18842}
}

@article{enomoto2026readmore,
  title   = {Read More, Think More: Revisiting Observation Reduction for Web Agents},
  author  = {Enomoto, Masafumi and Obara, Ryoma and Zhang, Haochen and Oyamada, Masafumi},
  journal = {arXiv preprint arXiv:2604.01535},
  year    = {2026},
  url     = {https://arxiv.org/abs/2604.01535}
}

@inproceedings{lin2025showui,
  title     = {ShowUI: One Vision-Language-Action Model for GUI Visual Agent},
  author    = {Lin, Kevin Qinghong and Li, Linjie and Gao, Difei and Yang, Zhengyuan and Wu, Shiwei and Bai, Zechen and Lei, Weixian and Wang, Lijuan and Shou, Mike Zheng},
  booktitle = {Proceedings of the IEEE/CVF Conference on Computer Vision and Pattern Recognition},
  year      = {2025},
  pages     = {19498--19508},
  doi       = {10.1109/CVPR52734.2025.01816},
  url       = {https://openaccess.thecvf.com/content/CVPR2025/html/Lin_ShowUI_One_Vision-Language-Action_Model_for_GUI_Visual_Agent_CVPR_2025_paper.html}
}

@article{li2025screenspotpro,
  title   = {ScreenSpot-Pro: GUI Grounding for Professional High-Resolution Computer Use},
  author  = {Li, Kaixin and Meng, Ziyang and Lin, Hongzhan and Luo, Ziyang and Tian, Yuchen and Ma, Jing and Huang, Zhiyong and Chua, Tat-Seng},
  journal = {arXiv preprint arXiv:2504.07981},
  year    = {2025},
  url     = {https://arxiv.org/abs/2504.07981}
}

@inproceedings{chen2025lessismore,
  title     = {Less is More: Empowering GUI Agent with Context-Aware Simplification},
  author    = {Chen, Gongwei and Zhou, Xurui and Shao, Rui and Lyu, Yibo and Zhou, Kaiwen and Wang, Shuai and Li, Wentao and Li, Yinchuan and Qi, Zhongang and Nie, Liqiang},
  booktitle = {Proceedings of the IEEE/CVF International Conference on Computer Vision},
  year      = {2025},
  pages     = {5901--5911},
  url       = {https://openaccess.thecvf.com/content/ICCV2025/html/Chen_Less_is_More_Empowering_GUI_Agent_with_Context-Aware_Simplification_ICCV_2025_paper.html}
}

@article{zhou2026colorbrowseragent,
  title   = {ColorBrowserAgent: Complex Long-Horizon Browser Agent with Adaptive Knowledge Evolution},
  author  = {Zhou, Jiamu and Wang, Jihong and Zhang, Weiming and Liu, Weiwen and Zhang, Zhuosheng and Lou, Xingyu and Zhang, Weinan and Deng, Huarong and Wang, Jun},
  journal = {arXiv preprint arXiv:2601.07262},
  year    = {2026},
  url     = {https://arxiv.org/abs/2601.07262}
}

@inproceedings{xie2025guiexplorer,
  title     = {GUI-explorer: Autonomous Exploration and Mining of Transition-aware Knowledge for GUI Agent},
  author    = {Xie, Bin and Shao, Rui and Chen, Gongwei and Zhou, Kaiwen and Li, Yinchuan and Liu, Jie and Zhang, Min and Nie, Liqiang},
  booktitle = {Proceedings of the 63rd Annual Meeting of the Association for Computational Linguistics},
  year      = {2025},
  url       = {https://aclanthology.org/2025.acl-long.282/}
}

@article{lee2025verisafe,
  title   = {VeriSafe Agent: Safeguarding Mobile GUI Agent via Logic-based Action Verification},
  author  = {Lee, Jungjae and Lee, Dongjae and Choi, Chihun and Im, Youngmin and Wi, Jaeyoung and Heo, Kihong and Oh, Sangeun and Lee, Sunjae and Shin, Insik},
  journal = {arXiv preprint arXiv:2503.18492},
  year    = {2025},
  url     = {https://arxiv.org/abs/2503.18492}
}

@article{wang2025mmbenchgui,
  title   = {MMBench-GUI: Hierarchical Multi-Platform Evaluation Framework for GUI Agents},
  author  = {Wang, Xuehui and Wu, Zhenyu and Xie, JingJing and Ding, Zichen and Yang, Bowen and Li, Zehao and Liu, Zhaoyang and Li, Qingyun and Dong, Xuan and Chen, Zhe and Wang, Weiyun and Zhao, Xiangyu and Chen, Jixuan and Duan, Haodong and Xie, Tianbao and Yang, Chenyu and Su, Shiqian and Yu, Yue and Huang, Yuan and Liu, Yiqian and Zhang, Xiao and Zhang, Yanting and Yue, Xiangyu and Su, Weijie and Zhu, Xizhou and Shen, Wei and Dai, Jifeng and Wang, Wenhai},
  journal = {arXiv preprint arXiv:2507.19478},
  year    = {2025},
  url     = {https://arxiv.org/abs/2507.19478}
}

@misc{lin2025mobilewizard,
  title        = {MobileWizard: A Data-Efficient GUI Agent with Structured Reasoning and Progressive Reinforcement Learning},
  author       = {Lin, Weifeng and Chai, Yuxiang and Xiao, Han and Bian, Liuyang and Liu, Guangyi and Liu, Liang and Ren, Shuai and Shi, Penggang and Wen, Yafei and Chen, Xiaoxin and Zhou, Aojun and Li, Hongsheng},
  year         = {2025},
  howpublished = {OpenReview},
  url          = {https://openreview.net/forum?id=Aobvdp3XmP}
}

@inproceedings{zhang2025agentcpmgui,
  title   = {AgentCPM-GUI: Building Mobile-Use Agents with Reinforcement Fine-Tuning},
  author  = {Zhang, Zhong and Lu, Yaxi and Fu, Yikun and Huo, Yupeng and Yang, Shenzhi and Wu, Yesai and Si, Han and Cong, Xin and Chen, Haotian and Lin, Yankai and Xie, Jie and Zhou, Wei and Xu, Wang and Zhang, Yuanheng and Su, Zhou and Zhai, Zhongwu and Liu, Xiaoming and Mei, Yudong and Xu, Jianming and Tian, Hongyan and Wang, Chongyi and Chen, Chi and Yao, Yuan and Liu, Zhiyuan and Sun, Maosong},
  booktitle = {Proceedings of the 2025 Conference on Empirical Methods in Natural Language Processing: System Demonstrations},
  year      = {2025},
  pages     = {155--180},
  doi       = {10.18653/v1/2025.emnlp-demos.12},
  url       = {https://aclanthology.org/2025.emnlp-demos.12/}
}

@article{abhyankar2025osworldhuman,
  title   = {OSWorld-Human: Benchmarking the Efficiency of Computer-Use Agents},
  author  = {Abhyankar, Reyna and Qi, Qi and Zhang, Yiying},
  journal = {arXiv preprint arXiv:2506.16042},
  year    = {2025},
  url     = {https://arxiv.org/abs/2506.16042}
}

@article{agashe2025agents2,
  title   = {Agent S2: A Compositional Generalist-Specialist Framework for Computer Use Agents},
  author  = {Agashe, Saaket and Wong, Kyle and Tu, Vincent and Yang, Jiachen and Li, Ang and Wang, Xin Eric},
  year    = {2025},
  journal = {arXiv preprint arXiv:2504.00906},
  doi     = {10.48550/arXiv.2504.00906},
  url     = {https://arxiv.org/abs/2504.00906}
}

@misc{song2025coact1,
  title   = {CoAct-1: Computer-using Multi-agent System with Coding Actions},
  author  = {Song, Linxin and Dai, Yutong and Prabhu, Viraj and Zhang, Jieyu and Shi, Taiwei and Li, Li and Li, Junnan and Savarese, Silvio and Chen, Zeyuan and Zhao, Jieyu and Xu, Ran and Xiong, Caiming},
  year    = {2026},
  howpublished = {OpenReview, ICLR 2026 poster},
  url     = {https://openreview.net/forum?id=l1MQVgIKEU}
}

@inproceedings{park2025rvlm,
  title={R-VLM: Region-Aware Vision Language Model for Precise GUI Grounding},
  author={Park, Joonhyung and Tang, Peng and Das, Sagnik and Appalaraju, Srikar and Singh, Kunwar Yashraj and Manmatha, R. and Ghadar, Shabnam},
  booktitle={Findings of the Association for Computational Linguistics: ACL 2025},
  year={2025},
  pages={9669--9685},
  doi={10.18653/v1/2025.findings-acl.501},
  url={https://aclanthology.org/2025.findings-acl.501/}
}

@inproceedings{wu2025dimogui,
  title={DiMo-GUI: Advancing Test-time Scaling in GUI Grounding via Modality-Aware Visual Reasoning},
  author={Wu, Hang and Chen, Hongkai and Cai, Yujun and Liu, Chang and Ye, Qingwen and Yang, Ming-Hsuan and Wang, Yiwei},
  booktitle={Proceedings of the 2025 Conference on Empirical Methods in Natural Language Processing},
  year={2025},
  pages={26246--26256},
  doi={10.18653/v1/2025.emnlp-main.1334},
  url={https://aclanthology.org/2025.emnlp-main.1334/}
}

@misc{wu2025guiactor,
  title={GUI-Actor: Coordinate-Free Visual Grounding for GUI Agents},
  author={Wu, Qianhui and Cheng, Kanzhi and Yang, Rui and Zhang, Chaoyun and Yang, Jianwei and Jiang, Huiqiang and Mu, Jian and Peng, Baolin and Qiao, Bo and Tan, Reuben and Qin, Si and Liden, Lars and Lin, Qingwei and Zhang, Huan and Zhang, Tong and Zhang, Jianbing and Zhang, Dongmei and Gao, Jianfeng},
  year={2025},
  howpublished={NeurIPS 2025 poster},
  doi={10.48550/arXiv.2506.03143},
  url={https://arxiv.org/abs/2506.03143}
}

@misc{liu2025guirise,
  title={GUI-Rise: Structured Reasoning and History Summarization for GUI Navigation},
  author={Liu, Tao and Wang, Chongyu and Li, Rongjie and Yu, Yingchen and He, Xuming and Bai, Song},
  year={2025},
  howpublished={OpenReview, NeurIPS 2025 poster},
  url={https://openreview.net/forum?id=YMPYLesItf}
}

@misc{yu2026polyskill,
  title={PolySkill: Learning Generalizable Skills Through Polymorphic Abstraction For Continual Learning},
  author={Yu, Simon and Li, Gang and Shi, Weiyan and Qi, Peng},
  year={2026},
  howpublished={OpenReview, ICLR 2026 poster},
  url={https://openreview.net/forum?id=KdEsujyiSV}
}

@misc{cai2026senseact,
  title={SenseAct: Structuring GUI Actions for Reliable Planning and Verification},
  author={Cai, Hongtian and Ma, Tianyi and Shao, Jie-Jing and Tang, Tianyi and Tsang, Ivor and Lyu, Yueming and Yin, Haiyan},
  year={2026},
  howpublished={OpenReview, ICLR 2026 AIWILD workshop},
  url={https://openreview.net/forum?id=0DznNQFW4g}
}

@misc{chen2025guishepherd,
  title={GUI-Shepherd: Reliable Process Reward and Verification for Long-Sequence GUI Tasks},
  author={Chen, Cong and Ji, Kaixiang and Zhong, Hao and Zhu, Muzhi and Li, Anzhou and Gan, Guo and Huang, Ziyuan and Zou, Cheng and Liu, Jiajia and Chen, Jingdong and Chen, Hao and Shen, Chunhua},
  year={2025},
  howpublished={OpenReview, submitted to ICLR 2026},
  url={https://openreview.net/forum?id=9hM4YRMhfT}
}

@misc{dihan2025weboperator,
  title={WebOperator: Action-Aware Tree Search for Autonomous Agents in Web Environment},
  author={Dihan, Mahir Labib and Hashem, Tanzima and Ali, Mohammed Eunus and Parvez, Md Rizwan},
  year={2025},
  howpublished={OpenReview, submitted to ICLR 2026},
  url={https://openreview.net/forum?id=vnEuxLVFmN}
}

@misc{zhou2025guig1,
  title={GUI-G1: Understanding R1-Zero-Like Training for Visual Grounding in GUI Agents},
  author={Zhou, Yuqi and Dai, Sunhao and Wang, Shuai and Zhou, Kaiwen and Jia, Qinglin and Xu, Jun},
  year={2025},
  howpublished={OpenReview, NeurIPS 2025 poster},
  url={https://openreview.net/forum?id=1XLjrmKZ4p}
}

@misc{chen2025adaguir1,
  title={Difficulty-Aware Reasoning for Mobile GUI Automation via Reinforcement Fine-Tuning},
  author={Chen, Jiafu and Lv, Rui and Jing, Hongyi and Dang, Ziqiang and Fang, Shuo and Ma, Chenguang and Zhao, Lei and Teng, Jiajie},
  year={2025},
  howpublished={OpenReview, submitted to ICLR 2026},
  url={https://openreview.net/forum?id=Ric2If6Xur}
}

@inproceedings{liu2026infiguiagent,
  title={InfiGUIAgent: A Multimodal Generalist GUI Agent with Native Reasoning and Reflection},
  author={Liu, Yuhang and Li, Pengxiang and Wei, Zishu and Xie, Congkai and Hu, Xueyu and Xu, Xinchen and Zhang, Shengyu and Han, Xiaotian and Yang, Hongxia and Wu, Fei},
  booktitle={Proceedings of the 19th Conference of the European Chapter of the Association for Computational Linguistics (Volume 1: Long Papers)},
  year={2026},
  pages={1035--1051},
  doi={10.18653/v1/2026.eacl-long.45},
  url={https://aclanthology.org/2026.eacl-long.45/}
}

@misc{lee2026intentcua,
  title={IntentCUA: Learning Intent-level Representations for Skill Abstraction and Multi-Agent Planning in Computer-Use Agents},
  author={Lee, Seoyoung and Yoon, Seobin and Lee, Seongbeen and Park, Dayoung and Kim, Doyeon and Chun, Yoojung and Sim, Joo Yong},
  year={2026},
  howpublished={OpenReview, AAMAS 2026 full paper},
  url={https://openreview.net/forum?id=qeSsHx2wIr}
}

@misc{yang2025ultracua,
  title={UltraCUA: Scaling Computer Use Agent through GUI and Programmatic Control},
  author={Yang, Yuhao and Yang, Zhen and Dou, Zi-Yi and Nguyen, Anh Tuan and Attia, Omar and Szot, Andrew and You, Keen and Feng, Michael and Ramrakhya, Ram and Toshev, Alexander T. and Huang, Chao and Yang, Yinfei and Gan, Zhe},
  year={2025},
  howpublished={OpenReview, ICLR 2026 desk rejected submission},
  url={https://openreview.net/forum?id=yVP2ldY9lw}
}

@inproceedings{kang2026longhorizonui,
  title     = {LongHorizonUI: A Unified Framework for Robust Long-Horizon Task Automation of GUI Agent},
  author    = {Kang, Bin and Wen, Shaoguo and Bi, Yifei and Wu, Shunlong and Yuan, Xinbin and Shao, Rui and Wang, Junle and Tian, Zhuotao},
  booktitle = {The Thirteenth International Conference on Learning Representations},
  year      = {2026},
  url       = {https://openreview.net/forum?id=BK7Mk5d4WE}
}

@article{yang2026ossymphony,
  title   = {OS-Symphony: A Holistic Framework for Robust and Generalist Computer-Using Agent},
  author  = {Yang, Bowen and Jin, Kaiming and Wu, Zhenyu and Liu, Zhaoyang and Sun, Qiushi and Li, Zehao and Xie, JingJing and Liu, Zhoumianze and Xu, Fangzhi and Cheng, Kanzhi and Li, Qingyun and Wang, Yian and Qiao, Yu and Wang, Zun and Ding, Zichen},
  journal = {arXiv preprint arXiv:2601.07779},
  year    = {2026},
  url     = {https://arxiv.org/abs/2601.07779}
}

@article{mind2web,
  title   = {{Mind2Web}: Towards a Generalist Agent for the Web},
  author  = {Xiang Deng and Yu Gu and Boyuan Zheng and Shijie Chen and Samuel Stevens and Boshi Wang and Huan Sun and Yu Su},
  journal = {arXiv preprint arXiv:2306.06070},
  year    = {2023},
  doi     = {10.48550/arXiv.2306.06070},
  url     = {https://arxiv.org/abs/2306.06070}
}

@article{webarena,
  title   = {{WebArena}: A Realistic Web Environment for Building Autonomous Agents},
  author  = {Shuyan Zhou and Frank F. Xu and Hao Zhu and Xuhui Zhou and Robert Lo and Abishek Sridhar and Xianyi Cheng and Tianyue Ou and Yonatan Bisk and Daniel Fried and Uri Alon and Graham Neubig},
  journal = {arXiv preprint arXiv:2307.13854},
  year    = {2023},
  doi     = {10.48550/arXiv.2307.13854},
  url     = {https://arxiv.org/abs/2307.13854}
}

@article{visualwebarena,
  title   = {{VisualWebArena}: Evaluating Multimodal Agents on Realistic Visual Web Tasks},
  author  = {Jing Yu Koh and Robert Lo and Lawrence Jang and Vikram Duvvur and Ming Chong Lim and Po-Yu Huang and Graham Neubig and Shuyan Zhou and Ruslan Salakhutdinov and Daniel Fried},
  journal = {arXiv preprint arXiv:2401.13649},
  year    = {2024},
  doi     = {10.48550/arXiv.2401.13649},
  url     = {https://arxiv.org/abs/2401.13649}
}

@article{browsergym,
  title   = {The {BrowserGym} Ecosystem for Web Agent Research},
  author  = {{Le Sellier De Chezelles}, Thibault and Maxime Gasse and Alexandre Drouin and Massimo Caccia and L{\'e}o Boisvert and Megh Thakkar and Tom Marty and Rim Assouel and Sahar Omidi Shayegan and Lawrence Keunho Jang and Xing Han L{\`u} and Ori Yoran and Dehan Kong and Frank F. Xu and Siva Reddy and Quentin Cappart and Graham Neubig and Ruslan Salakhutdinov and Nicolas Chapados and Alexandre Lacoste},
  journal = {arXiv preprint arXiv:2412.05467},
  year    = {2024},
  doi     = {10.48550/arXiv.2412.05467},
  url     = {https://arxiv.org/abs/2412.05467}
}

@article{osworld,
  title   = {{OSWorld}: Benchmarking Multimodal Agents for Open-Ended Tasks in Real Computer Environments},
  author  = {Tianbao Xie and Danyang Zhang and Jixuan Chen and Xiaochuan Li and Siheng Zhao and Ruisheng Cao and Toh Jing Hua and Zhoujun Cheng and Dongchan Shin and Fangyu Lei and Yitao Liu and Yiheng Xu and Shuyan Zhou and Silvio Savarese and Caiming Xiong and Victor Zhong and Tao Yu},
  journal = {arXiv preprint arXiv:2404.07972},
  year    = {2024},
  doi     = {10.48550/arXiv.2404.07972},
  url     = {https://arxiv.org/abs/2404.07972}
}

@article{windowsagentarena,
  title   = {{Windows Agent Arena}: Evaluating Multi-Modal OS Agents at Scale},
  author  = {Rogerio Bonatti and Dan Zhao and Francesco Bonacci and Dillon Dupont and Sara Abdali and Yinheng Li and Yadong Lu and Justin Wagle and Kazuhito Koishida and Arthur Bucker and Lawrence Jang and Zack Hui},
  journal = {arXiv preprint arXiv:2409.08264},
  year    = {2024},
  doi     = {10.48550/arXiv.2409.08264},
  url     = {https://arxiv.org/abs/2409.08264}
}

@article{efficientmllmsurvey,
  title   = {Efficient Multimodal Large Language Models: A Survey},
  author  = {Yizhang Jin and Jian Li and Yexin Liu and Tianjun Gu and Kai Wu and Zhengkai Jiang and Muyang He and Bo Zhao and Xin Tan and Zhenye Gan and Yabiao Wang and Chengjie Wang and Lizhuang Ma},
  journal = {arXiv preprint arXiv:2405.10739},
  year    = {2024},
  doi     = {10.48550/arXiv.2405.10739},
  url     = {https://arxiv.org/abs/2405.10739}
}

@article{efficientagentssurvey,
  title   = {Toward Efficient Agents: Memory, Tool learning, and Planning},
  author  = {Xiaofang Yang and Lijun Li and Heng Zhou and Tong Zhu and Xiaoye Qu and Yuchen Fan and Qianshan Wei and Rui Ye and Li Kang and Yiran Qin and Zhiqiang Kou and Daizong Liu and Qi Li and Ning Ding and Siheng Chen and Jing Shao},
  journal = {arXiv preprint arXiv:2601.14192},
  year    = {2026},
  doi     = {10.48550/arXiv.2601.14192},
  url     = {https://arxiv.org/abs/2601.14192}
}

@article{guiagentssurvey,
  title   = {{GUI} Agents: A Survey},
  author  = {Dang Nguyen and Jian Chen and Yu Wang and Gang Wu and Namyong Park and Zhengmian Hu and Hanjia Lyu and Junda Wu and Ryan Aponte and Yu Xia and Xintong Li and Jing Shi and Hongjie Chen and Viet Dac Lai and Zhouhang Xie and Sungchul Kim and Ruiyi Zhang and Tong Yu and Mehrab Tanjim and Nesreen K. Ahmed and Puneet Mathur and Seunghyun Yoon and Lina Yao and Branislav Kveton and Jihyung Kil and Thien Huu Nguyen and Trung Bui and Tianyi Zhou and Ryan A. Rossi and Franck Dernoncourt},
  journal = {arXiv preprint arXiv:2412.13501},
  year    = {2024},
  doi     = {10.48550/arXiv.2412.13501},
  url     = {https://arxiv.org/abs/2412.13501}
}

@article{computerusesurvey,
  title   = {A Comprehensive Survey of Agents for Computer Use: Foundations, Challenges, and Future Directions},
  author  = {Pascal J. Sager and Benjamin Meyer and Peng Yan and Rebekka von Wartburg-Kottler and Layan Etaiwi and Aref Enayati and Gabriel Nobel and Ahmed Abdulkadir and Benjamin F. Grewe and Thilo Stadelmann},
  journal = {arXiv preprint arXiv:2501.16150},
  year    = {2025},
  doi     = {10.48550/arXiv.2501.16150},
  url     = {https://arxiv.org/abs/2501.16150}
}

\newpage
\appendix
\onecolumn

\section{GUI-agent}
\label{sec:gui-agent-form}
A single next-action equation hides the fact that deployment cost accumulates over a full interaction. We therefore also view a GUI-agent episode as a system-level trajectory. Let $s_t$ denote the latent environment state, $o_t$ the GUI observation extracted from that state, $m_t$ the retained context or memory, $a_t$ the executed interface action, and $v_t$ a verification or reflection signal after execution. A compact abstraction is
\begin{align}
o_t &= \Omega(s_t), \label{eq:obs_model}\\
m_t &= U_m(m_{t-1}, o_t, a_{t-1}, v_{t-1}), \label{eq:memory_update}\\
a_t &\sim \pi_\theta(g, o_t, m_t), \label{eq:policy}\\
s_{t+1} &\sim \mathcal{T}(s_t, a_t), \label{eq:transition}\\
v_t &= V(g, s_t, o_t, a_t, s_{t+1}), \label{eq:verification}
\end{align}
where $\Omega$ covers screenshot, DOM, HTML, AxTree, parser, or hybrid perception modules; $U_m$ covers history summarization, retrieval, compression, or KV-cache reuse; $\pi_\theta$ covers the planner, grounding model, or orchestrator; $\mathcal{T}$ is the GUI environment transition; and $V$ denotes explicit verifiers, self-reflection, post-condition checks, or implicit task-progress feedback. The full execution is
\begin{equation}
\tau = (s_0, o_0, m_0, a_0, v_0, \ldots, s_T, o_T, m_T),
\end{equation}
with task outcome $S(\tau) \in \{0,1\}$ or a graded completion score.
Sections~\ref{sec:observation_efficiency}--\ref{sec:planner_system_efficiency} can be read as four ways to reduce this trajectory cost: cheaper observations, cheaper memory and context updates, fewer or more recoverable actions, and lower planner/runtime overhead.

For ease of reading in the main text, we collect the tree-structured taxonomy overview and the section-wise paper summary tables in this appendix.
\clearpage
\section{Literature Expansion Protocol}
\label{sec:appendix_protocol}

We expanded the seed draft subsection by subsection rather than treating literature search as a single flat retrieval problem. For each existing subsection, we began with the papers already cited in the draft, then used targeted search over arXiv, ACL Anthology, and OpenReview together with backward and forward citation chaining to surface closely related work. We retained papers only when they directly addressed efficiency in the relevant subsection: observation length or actionability, memory size or retrieval cost, action count or recoverability, or planner/runtime overhead. We excluded generic GUI grounding papers, pure benchmark-generation papers, and unrelated long-horizon agent work unless they directly informed the subsection's efficiency mechanism. Where a paper had both a preprint and a later venue version, we normalized it to a single canonical entry.

\section{Added Works by Subsection}
\label{sec:appendix_added_works}

\begin{table}[h]
\centering
\small
\caption{Additional works integrated beyond the seed draft, grouped by the existing subsection taxonomy.}
\label{tab:added_works}
\begin{tabularx}{\linewidth}{p{0.33\linewidth}X}
\toprule
Subsection & Added or normalized works \\
\midrule
Textual Observation Reduction & Searched adjacent pruning, retrieval, and adaptive formatting work; retained the seed pruning/retrieval cluster as the representative core because no equally central stable additions survived screening. \\
Region-Focused Visual Perception & Added R-VLM and DiMo-GUI to broaden the zoom, region-proposal, and modality-aware token-allocation line. \\
Observation Enrichment via Parsing and Hybridization & Added Read Anywhere Pointed / Tree-of-Lens and GUI-Actor; normalized Aria-UI to its ACL Findings 2025 version. \\
Summary-Based Compression and Selective Look-Back & Added GUI-Rise and retained PAL-UI, HiconAgent, and SimpAgent as complementary evidence on selective retrieval and history compression. \\
Runtime-Level Representation Compression & Retained the GUI-KV, ST-Lite, continuous-memory, and SecAgent cluster after screening for GUI-specific serving-level compression. \\
Action Abstraction & Added PolySkill and normalized CoAct-1 as a stable OpenReview publication signal for hybrid GUI-plus-code execution. \\
Action Pruning, Verification, and Recovery & Normalized BacktrackAgent to EMNLP 2025 and added GUI-Shepherd plus SenseAct as explicitly marked emerging evidence. \\
Exploration Control & Added WebOperator as a clearly labeled prepublication signal on search-aware, backtracking-heavy web control. \\
Planner-Side Efficiency & Added GUI-G1, AdaGUI-R1, and InfiGUIAgent; normalized MobileUse and AgentCPM-GUI status. \\
System Efficiency Beyond the Planner & Added IntentCUA, normalized Agent S2 and CoAct-1, and kept UltraCUA only in the challenges section because its current status is unstable. \\
\bottomrule
\end{tabularx}
\end{table}

\section{Taxonomy Figure and Section Summary Tables}
\label{sec:section-wise-table}

For ease of reading in the main text, we collect the tree-structured taxonomy overview and the section-wise paper summary tables in this appendix. \revgreen{In Tables~\ref{tab:obs_methods}--\ref{tab:planner_system_methods}, \emph{NR} means that the source paper does not report a comparable quantitative efficiency metric. Reported percentages, speedups, token counts, costs, and step counts are copied from the authors' baseline comparisons and should be read as an evidence ledger rather than a direct meta-analysis.}

%

\begingroup
\scriptsize
\setlength{\tabcolsep}{1.0pt}
\renewcommand{\arraystretch}{1.05}
\begin{longtable}{L{2.95cm} L{1.28cm} L{1.70cm} L{1.65cm} L{3.90cm} L{3.05cm} L{0.95cm}}
\caption{Observation-efficiency papers discussed in this survey, grouped by the existing subsection structure.}
\label{tab:obs_methods}\\
\toprule
Paper & Venue & Platform+UI & Core mechanism & \revgreen{Reported efficiency signal} & \revgreen{Unreported / shifted cost} & Code \\
\midrule
\endfirsthead
\toprule
Paper & Venue & Platform+UI & Core mechanism & \revgreen{Reported efficiency signal} & \revgreen{Unreported / shifted cost} & Code \\
\midrule
\endhead
\bottomrule
\endlastfoot
\tablegroup{Textual Observation Reduction}
Agent-E \cite{abuelsaad2024agente} & arXiv'24 & Web/DOM & DOM filtering & \revgreen{\textbf{150--220 s/task}; about \textbf{25 LLM calls/task}; raw DOM up to \textbf{800k tokens}.} & \revgreen{Filtered-token saving and DOM-filter latency are \textbf{NR}.} & \codelink{https://github.com/EmergenceAI/Agent-E} \\
Beyond Pixels \cite{schiepanski2025beyondpixels} & arXiv'25 & Web/DOM & DOM compression & \revgreen{Compressed DOM snapshots stay around the \textbf{$10^3$ token} order.} & \revgreen{DOM downsampling runtime is \textbf{NR}.} & \na \\
LineRetriever \cite{kerboua2025lineretriever} & arXiv'25 & Web/AxTree & line retrieval & \revgreen{Observation reduced by \textbf{61\%}/\textbf{72\%}/\textbf{73\%}; retrieves up to \textbf{10} chunks of \textbf{100} tokens.} & \revgreen{Retriever latency and extra model cost are \textbf{NR}.} & \na \\
FocusAgent \cite{kerboua2025focusagent} & arXiv'25 & Web/AxTree & AxTree filtering & \revgreen{\textbf{$>$50\%} average reduction, often \textbf{$>$80\%}; caps context at \textbf{2k tokens}.} & \revgreen{Full wall-clock retrieval overhead is \textbf{NR}.} & \na \\
Prune4Web \cite{zhang2025prune4web} & arXiv'25 & Web/DOM & DOM/action pruning & \revgreen{DOMs: \textbf{10k--100k tokens}; example shrinks \textbf{$>$500} elements to \textbf{$<$20}.} & \revgreen{Filter-program generation/runtime is \textbf{NR}.} & \na \\
Read More, Think More \cite{enomoto2026readmore} & arXiv'26 & Web/HTML+Ax & adaptive view & \revgreen{WorkArena HTML pages: \textbf{40k--500k tokens}; look-back settings: \textbf{4}/\textbf{9} steps.} & \revgreen{Adaptive-view saving and latency are \textbf{NR}.} & \na \\
\midrule
\tablegroup{Region-Focused Visual Perception}
SeeClick \cite{cheng2024seeclick} & ACL'24 & GUI/screen & GUI pretraining & \revgreen{\textbf{NR}; no reported token, latency, memory, or step saving.} & \revgreen{Screenshot-only runtime cost is \textbf{NR}.} & \codelink{https://github.com/njucckevin/SeeClick} \\
Ferret-UI \cite{you2024ferretui} & arXiv'24 & Mobile/screen & screen crops & \revgreen{Any-resolution processing splits each screen into \textbf{2 sub-images}.} & \revgreen{Crop encoding and inference overhead are \textbf{NR}.} & \na \\
R-VLM \cite{park2025rvlm} & ACL F.'25 & GUI/screen & region proposals & \revgreen{Reports about \textbf{5.6 s/sample} and up to \textbf{2$\times$} inference-latency cost.} & \revgreen{Region proposal shifts cost to extra inference.} & \na \\
RegionFocus \cite{luo2025regionfocus} & arXiv'25 & GUI/screen & zoom search & \revgreen{Average trajectory overhead \textbf{66.8\%}; step count increases by \textbf{19.74\%}.} & \revgreen{Zoom-search latency is only partially reported.} & \na \\
ShowUI \cite{lin2025showui} & CVPR'25 & GUI/screen & token pruning & \revgreen{Removes \textbf{33\%} redundant visual tokens; reports \textbf{1.4$\times$} training speedup.} & \revgreen{End-to-end agent latency is \textbf{NR}.} & \codelink{https://github.com/showlab/ShowUI} \\
DiMo-GUI \cite{wu2025dimogui} & EMNLP'25 & GUI/screen & adaptive zoom & \revgreen{Uses up to \textbf{7} zoom iterations.} & \revgreen{Extra zoom-call latency is \textbf{NR}.} & \na \\
ScreenSpot-Pro \cite{li2025screenspotpro} & arXiv'25 & GUI/high-res & cascaded search & \revgreen{Benchmark screens often exceed \textbf{3k$\times$2k} resolution.} & \revgreen{Benchmark only; runtime metric is \textbf{NR}.} & \na \\
SimpAgent \cite{chen2025lessismore} & ICCV'25 & GUI/screen+hist & screen masking & \revgreen{History image: \textbf{64 tokens}; action output: \textbf{10--20 tokens}; LLM branch: \textbf{27\%} FLOP reduction.} & \revgreen{Simplifier overhead and full-task latency are \textbf{NR}.} & \na \\
\midrule
\tablegroup{Observation Enrichment via Parsing and Hybridization}
Set-of-Mark \cite{yang2023som} & arXiv'23 & GUI/screen & region marks & \revgreen{\textbf{NR}; no comparable token, latency, or memory metric.} & \revgreen{Mark rendering cost is \textbf{NR}.} & \na \\
ScreenAI \cite{baechler2024screenai} & arXiv'24 & GUI/screen-text & screen captioning & \revgreen{Reports \textbf{670M}/\textbf{2B}/\textbf{5B} model scales and up to \textbf{812$^2$} input resolution.} & \revgreen{GUI-agent latency or token saving is \textbf{NR}.} & \na \\
OmniParser \cite{lu2024omniparser} & arXiv'24 & GUI/parser & screen parsing & \revgreen{Adds \textbf{2} parser models; one setting keeps top-\textbf{50} relevant elements.} & \revgreen{Parser latency and extra calls are \textbf{NR}.} & \codelink{https://github.com/microsoft/OmniParser} \\
Tree-of-Lens \cite{fan2024treeoflens} & arXiv'24 & GUI/pointed & layout tree & \revgreen{\textbf{NR}; no reported token, latency, or parser-cost saving.} & \revgreen{Detector/layout-tree runtime is \textbf{NR}.} & \codelink{https://github.com/eric-ai-lab/Screen-Point-and-Read} \\
GUI-Actor \cite{wu2025guiactor} & NeurIPS'25 & GUI/screen & region actions & \revgreen{Action head adds about \textbf{20M} params for 2B and \textbf{100M} for 7B; candidates in \textbf{one forward pass}.} & \revgreen{Wall-clock latency is \textbf{NR}.} & \codelink{https://github.com/microsoft/GUI-Actor} \\
Agent-S \cite{agashe2024agents} & arXiv'24 & OS/img+AX & hybrid perception & \revgreen{\textbf{NR}; no observation-side cost metric.} & \revgreen{Step count and wall-clock overhead are \textbf{NR}.} & \codelink{https://github.com/simular-ai/Agent-S} \\
Ferret-UI 2 \cite{li2024ferretui2} & arXiv'24 & Cross-platform & adaptive scaling & \revgreen{\textbf{NR}; no runtime or token-efficiency metric.} & \revgreen{Adaptive-scaling overhead is \textbf{NR}.} & \na \\
UGround \cite{gou2024uground} & ICLR'25 & GUI/screen & visual grounding & \revgreen{Adaptive resolution uses about \textbf{2/3} of fixed 1344$\times$1344 visual tokens.} & \revgreen{End-to-end agent latency is \textbf{NR}.} & \codelink{https://github.com/OSU-NLP-Group/UGround} \\
Aria-UI \cite{yang2025aria} & ACL F.'25 & GUI/screen & instruction grounding & \revgreen{\textbf{3.9B} activated params; phase-1 training: about \textbf{18 h}/\textbf{10k} steps.} & \revgreen{Exact inference latency is \textbf{NR}.} & \na \\
Aguvis \cite{xu2024aguvis} & arXiv'24 & GUI/screen & pure-vision loop & \revgreen{Text agents: \textbf{4k--6k tokens/step}; pure-vision loop: \textbf{1,196 tokens}; \textbf{70\%} input-token reduction.} & \revgreen{Vision-encoder latency is \textbf{NR}.} & \codelink{https://github.com/xlang-ai/aguvis} \\
UI-TARS \cite{qin2025uitars} & arXiv'25 & GUI/screen & native screen agent & \revgreen{Model scales: \textbf{2B}/\textbf{7B}/\textbf{72B}; training budget about \textbf{50B tokens}; OSWorld budgets: \textbf{15}/\textbf{50} steps.} & \revgreen{Training cost and inference latency are \textbf{NR}.} & \codelink{https://github.com/bytedance/UI-TARS} \\
\end{longtable}
\endgroup

\begingroup
\scriptsize
\setlength{\tabcolsep}{1.0pt}
\renewcommand{\arraystretch}{1.05}
\begin{longtable}{L{2.95cm} L{1.28cm} L{1.70cm} L{1.65cm} L{3.90cm} L{3.05cm} L{0.95cm}}
\caption{Context- and memory-efficiency papers discussed in this survey, grouped by the existing subsection structure.}
\label{tab:memory_methods}\\
\toprule
Paper & Venue & Platform+UI & Core mechanism & \revgreen{Reported efficiency signal} & \revgreen{Unreported / shifted cost} & Code \\
\midrule
\endfirsthead
\toprule
Paper & Venue & Platform+UI & Core mechanism & \revgreen{Reported efficiency signal} & \revgreen{Unreported / shifted cost} & Code \\
\midrule
\endhead
\bottomrule
\endlastfoot
\tablegroup{Summary-Based Compression and Selective Look-Back}
Agent-S \cite{agashe2024agents} & arXiv'24 & OS/img+AX & episodic memory & \revgreen{\textbf{NR}; no memory footprint, prompt-growth, or latency metric.} & \revgreen{Memory update/storage overhead is \textbf{NR}.} & \codelink{https://github.com/simular-ai/Agent-S} \\
ColorBrowserAgent \cite{zhou2026colorbrowseragent} & arXiv'26 & Web/GUI & progress summaries & \revgreen{Interaction horizon capped at \textbf{30 steps}.} & \revgreen{Summary-token saving and latency are \textbf{NR}.} & \na \\
GUI-Rise \cite{liu2025guirise} & NeurIPS'25 & GUI/img & task summaries & \revgreen{\textbf{NR}; no summary compression ratio or retrieval-cost metric.} & \revgreen{Summary and retrieval overhead are \textbf{NR}.} & \codelink{https://github.com/Leon022/GUI-Rise-code} \\
PAL-UI \cite{liu2025palui} & arXiv'25 & GUI/img & summary + retrieval & \revgreen{\textbf{NR}; no memory footprint or latency metric.} & \revgreen{Retrieval cost is \textbf{NR}.} & \na \\
HiconAgent \cite{zhou2025hiconagent} & arXiv'25 & GUI/img & context sampling & \revgreen{Compressed history uses \textbf{25.21T FLOPs} vs. \textbf{35.75T} uncompressed 3B and \textbf{62.31T} 7B; reports \textbf{60\%} FLOP reduction.} & \revgreen{Wall-clock latency and measured memory footprint are \textbf{NR}.} & \codelink{https://github.com/iLearn-Lab/CVPR26-HiconAgent} \\
SimpAgent \cite{chen2025lessismore} & ICCV'25 & GUI/img+hist & history pruning & \revgreen{History image: \textbf{64 tokens}; LLM branch: \textbf{27\%} FLOP reduction.} & \revgreen{Simplifier overhead and full-task latency are \textbf{NR}.} & \na \\
Read More, Think More \cite{enomoto2026readmore} & arXiv'26 & Web/history & history diffs & \revgreen{HTML pages: \textbf{40k--500k tokens}; history settings: \textbf{4}/\textbf{9} past steps.} & \revgreen{History-diff compression ratio and latency are \textbf{NR}.} & \na \\
\midrule
\tablegroup{Runtime-Level Representation Compression}
GUI-KV \cite{huang2025guikv} & arXiv'25 & GUI/VLM cache & KV pruning & \revgreen{\textbf{38.9\%} fewer MFLOPs/decoded token at 5 screenshots; \textbf{5--20\%} cache budgets; 5 screenshots can exceed \textbf{80GB} GPU memory.} & \revgreen{Prefill and end-to-end latency remain partially unresolved.} & \na \\
ST-Lite \cite{zhou2026stlite} & arXiv'26 & GUI/VLM cache & cache pruning & \revgreen{\textbf{10--20\%} cache budgets; \textbf{2.45$\times$} decoding speedup; \textbf{1.40$\times$} end-to-end speedup; prefill about \textbf{1.0$\times$}.} & \revgreen{Speedup is mostly decode-side.} & \na \\
Continuous Memory \cite{wu2025continuousmemory} & arXiv'25 & GUI/latent mem. & latent memory & \revgreen{Each trajectory becomes \textbf{8 embeddings}; raw trajectories can exceed \textbf{15k tokens}; collection cost about \textbf{\$4k}; tunes \textbf{1.2\%} params.} & \revgreen{Latent retrieval and embedding overhead are \textbf{NR}.} & \codelink{https://github.com/WenyiWU0111/CoMEM-Agent} \\
SecAgent \cite{xie2026secagent} & arXiv'26 & Mobile/sem. ctx & semantic summaries & \revgreen{N=1 to N=5: ITC \textbf{2,239$\rightarrow$3,642} (+\textbf{62.7\%}); TTFT \textbf{0.11$\rightarrow$0.22}; TPS \textbf{140$\rightarrow$122}.} & \revgreen{Full-task runtime and training cost are not unified.} & \na \\
\end{longtable}
\endgroup

\begingroup
\scriptsize
\setlength{\tabcolsep}{1.0pt}
\renewcommand{\arraystretch}{1.05}
\begin{longtable}{L{2.95cm} L{1.28cm} L{1.70cm} L{1.65cm} L{3.90cm} L{3.05cm} L{0.95cm}}
\caption{Action-efficiency papers discussed in this survey, grouped by the existing subsection structure.}
\label{tab:action_methods}\\
\toprule
Paper & Venue & Platform+UI & Core mechanism & \revgreen{Reported efficiency signal} & \revgreen{Unreported / shifted cost} & Code \\
\midrule
\endfirsthead
\toprule
Paper & Venue & Platform+UI & Core mechanism & \revgreen{Reported efficiency signal} & \revgreen{Unreported / shifted cost} & Code \\
\midrule
\endhead
\bottomrule
\endlastfoot
\tablegroup{Action Abstraction}
SkillWeaver \cite{zheng2025skillweaver} & arXiv'25 & Web/APIs & skill APIs & \revgreen{\textbf{NR}; no reused-skill action-count or latency saving.} & \revgreen{API overhead is \textbf{NR}.} & \codelink{https://github.com/OSU-NLP-Group/SkillWeaver} \\
PolySkill \cite{yu2026polyskill} & ICLR'26 & Web/skills & polymorphic skills & \revgreen{Learned functions typically cover \textbf{2--5 GUI steps}.} & \revgreen{Exact step reduction and skill-search cost are \textbf{NR}.} & \na \\
Mobile-Agent-E \cite{wang2025mobileagente} & arXiv'25 & Mobile/skills & shortcut skills & \revgreen{\textbf{NR}; no shortcut action-count or latency saving.} & \revgreen{Shortcut discovery/runtime cost is \textbf{NR}.} & \codelink{https://github.com/X-PLUG/MobileAgent} \\
ActionEngine \cite{zhong2026actionengine} & arXiv'26 & GUI/programs & compiled routines & \revgreen{Cost \textbf{\$0.71$\rightarrow$\$0.06} (\textbf{11.83$\times$}); latency \textbf{237.5$\rightarrow$118.3 s}; input \textbf{62.3k$\rightarrow$8.1k tokens}; calls \textbf{10.2$\rightarrow$1.8}.} & \revgreen{Compiler, validator, and first-recovery latency remain shifted costs.} & \na \\
CoAct-1 \cite{song2025coact1} & ICLR'26 & OS/GUI+code & code execution & \revgreen{Average steps: \textbf{10.15} vs. GTA-1 \textbf{15.22} and UI-TARS \textbf{14.90}.} & \revgreen{Code-execution overhead and sandboxing cost are \textbf{NR}.} & \na \\
\midrule
\tablegroup{Action Pruning, Verification, and Recovery}
Prune4Web \cite{zhang2025prune4web} & arXiv'25 & Web/DOM & candidate pruning & \revgreen{Example candidate set shrinks from \textbf{$>$500} DOM elements to \textbf{$<$20}.} & \revgreen{Pruner generation/runtime latency is \textbf{NR}.} & \na \\
V-Droid \cite{dai2025vdroid} & arXiv'25 & Mobile/img & action scoring & \revgreen{Input \textbf{2.6k--8.9k tokens}; \textbf{0.7 s/decision}; \textbf{4.3 s/step}; typical agents reported above \textbf{20 s/step}.} & \revgreen{Device/controller overhead is not separated.} & \na \\
VeriSafe Agent \cite{lee2025verisafe} & arXiv'25 & Mobile/img & intent checks & \revgreen{Recovery examples need \textbf{1--3 actions}; tasks above \textbf{10 steps} can exceed \textbf{\$1}.} & \revgreen{Verifier latency and cost curves are not tabulated.} & \na \\
GUI-Shepherd \cite{chen2025guishepherd} & OpenRev'25 & GUI/process & reward verifier & \revgreen{\textbf{NR}; no verifier-call, latency, or cost metric.} & \revgreen{Extra verifier calls and wall-clock latency are \textbf{NR}.} & \na \\
SenseAct \cite{cai2026senseact} & ICLR Wk.'26 & GUI/typed acts & typed checks & \revgreen{UI exposure reduced by \textbf{65.51\%}.} & \revgreen{Typed-check and recovery latency are \textbf{NR}.} & \na \\
BacktrackAgent \cite{wu2025backtrackagent} & EMNLP'25 & Mobile/img & backtracking & \revgreen{Speed ratios \textbf{0.451}/\textbf{0.517}/\textbf{0.482}; action execution about \textbf{0.25 s}.} & \revgreen{Recovery shifts cost to slower execution.} & \na \\
LongHorizonUI \cite{kang2026longhorizonui} & ICLR'26 & OS/img+idx & rollback loop & \revgreen{Average trajectory \textbf{24.6 steps}, max \textbf{37}; verifier choice changes latency \textbf{8.26 s$\rightarrow$5.74--6.59 s/step}.} & \revgreen{Rollback verifier-call overhead remains.} & \na \\
\midrule
\tablegroup{Exploration Control}
LASER \cite{ma2023laser} & arXiv'23 & Web/text & state search & \revgreen{\textbf{NR}; no search-cost, latency, or action-count saving.} & \revgreen{Exploration overhead is \textbf{NR}.} & \na \\
Auto-Intent \cite{kim2024autointent} & arXiv'24 & Web/demo-guided & intent guidance & \revgreen{Training uses about \textbf{1k A100-40GB GPU hours} plus \textbf{0.4k} GPU-hour search; inference up to \textbf{5 tokens}.} & \revgreen{Action-count saving is \textbf{NR}.} & \na \\
OpenWebVoyager \cite{he2024openwebvoyager} & arXiv'24 & Web/mm & explore-optimize & \revgreen{Samples each task trajectory up to \textbf{5} times; keeps at most \textbf{3 screenshots} in context.} & \revgreen{Exploration amortization, latency, and cost are \textbf{NR}.} & \na \\
GUI-explorer \cite{xie2025guiexplorer} & ACL'25 & GUI/mixed & transition mining & \revgreen{\textbf{66 s} per interaction step; ranker \textbf{28.5 s}; mining is \textbf{42.9\%} of runtime.} & \revgreen{Transition mining is a major overhead.} & \na \\
WebOperator \cite{dihan2025weboperator} & OpenRev'25 & Web/search & tree search & \revgreen{Tree search uses \textbf{31.34 actions} vs. \textbf{24.79} naive (+\textbf{6.55}); up to \textbf{3} candidates/step.} & \revgreen{Backtracking increases action count.} & \na \\
MobileUse \cite{li2025mobileuse} & OpenRev'25 & Mobile/img & proactive explore & \revgreen{Exploration cap: \textbf{100 steps/app} at \textbf{19.5 s/step}; reflection overhead reducible to \textbf{10\%}.} & \revgreen{Exploration cost is amortized, not eliminated.} & \na \\
\end{longtable}
\endgroup

\begingroup
\scriptsize
\setlength{\tabcolsep}{1.0pt}
\renewcommand{\arraystretch}{1.05}
\begin{longtable}{L{2.95cm} L{1.28cm} L{1.70cm} L{1.65cm} L{3.90cm} L{3.05cm} L{0.95cm}}
\caption{Planner-side and system-efficiency papers discussed in this survey, grouped by the existing subsection structure.}
\label{tab:planner_system_methods}\\
\toprule
Paper & Venue & Platform+UI & Core mechanism & \revgreen{Reported efficiency signal} & \revgreen{Unreported / shifted cost} & Code \\
\midrule
\endfirsthead
\toprule
Paper & Venue & Platform+UI & Core mechanism & \revgreen{Reported efficiency signal} & \revgreen{Unreported / shifted cost} & Code \\
\midrule
\endhead
\bottomrule
\endlastfoot
\tablegroup{Planner-Side Efficiency}
AndroidWorld \cite{rawles2024androidworld} & arXiv'24 & Mobile/eval & realistic eval & \revgreen{Benchmark footprint about \textbf{2GB memory} and \textbf{8GB disk}; \textbf{116} tasks across \textbf{20} apps.} & \revgreen{Benchmark only; agent latency/cost metric is \textbf{NR}.} & \na \\
MMBench-GUI \cite{wang2025mmbenchgui} & arXiv'25 & Cross-platform eval & efficiency eval & \revgreen{\textbf{50-step} budget; redundant-step cost EQ2 is \textbf{7--8}; privacy noise \textbf{40\%}; specialist cost \textbf{16\%}.} & \revgreen{Evaluation metrics expose cost but do not optimize runtime.} & \na \\
OSWorld-Human \cite{abhyankar2025osworldhuman} & arXiv'25 & OS/human eval & human profiling & \revgreen{Agents need \textbf{2.7$\times$--4.3$\times$} more steps; reflection takes \textbf{76\%--96\%} of task latency.} & \revgreen{Profiling only; no new optimization.} & \na \\
UI-R1 \cite{lu2025uir1} & arXiv'25 & GUI action pred. & RL policy & \revgreen{RFT uses only \textbf{136} training samples; training takes about \textbf{8 h} on \textbf{8 RTX 4090} GPUs.} & \revgreen{Inference latency and output-token cost are \textbf{NR}; high pixel settings can cause OOM.} & \na \\
Think Twice, Click Once \cite{tang2025thinktwice} & arXiv'25 & GUI grounding & adaptive reasoning & \revgreen{Adaptive slow thinking raises processing time \textbf{2.6$\rightarrow$5.4 s}; data include \textbf{300k} examples and \textbf{150k} slow samples.} & \revgreen{Higher reasoning depth trades for about \textbf{2$\times$} latency.} & \na \\
GUI-G1 \cite{zhou2025guig1} & OpenRev'25 & GUI grounding & fast-thinking RL & \revgreen{Output tokens: \textbf{37/39/39} on mobile/desktop/web vs. \textbf{107/107/114} for InfiGUI-R1, about \textbf{1/3}.} & \revgreen{Wall-clock latency is \textbf{NR}.} & \na \\
AdaGUI-R1 \cite{chen2025adaguir1} & OpenRev'25 & Mobile/GUI & reasoning schedule & \revgreen{Unnecessary reasoning tokens reduced by \textbf{40\%}; harder-example schedule adds \textbf{23.5\%} FLOPs.} & \revgreen{Latency and end-to-end cost are \textbf{NR}.} & \na \\
MobileUse \cite{li2025mobileuse} & OpenRev'25 & Mobile/img & on-demand reflection & \revgreen{Reflection overhead can be reduced to \textbf{10\%}; exploration costs up to \textbf{100 steps/app} at \textbf{19.5 s/step}.} & \revgreen{Exploration cost is amortized across tasks.} & \na \\
MobileWizard \cite{lin2025mobilewizard} & OpenRev'25 & Mobile/img & structured planning & \revgreen{Uses \textbf{24.5k} public trajectories plus \textbf{300} remedial trajectories; fewer than \textbf{50k} trajectories.} & \revgreen{Planner runtime, latency, and memory cost are \textbf{NR}.} & \na \\
AgentCPM-GUI \cite{zhang2025agentcpmgui} & EMNLP SD.'25 & Mobile/img & compact actions & \revgreen{\textbf{470k} atomic steps, about \textbf{8.5 steps/trajectory}; retains last \textbf{4} actions/images; max\_new\_tokens \textbf{2048}.} & \revgreen{Actual output-token saving and latency are \textbf{NR}.} & \codelink{https://github.com/OpenBMB/AgentCPM-GUI} \\
Agent S2 \cite{agashe2025agents2} & COLM'25 & OS/img+mods & planner split & \revgreen{\textbf{NR}; no specialist-routing latency, cost, or memory metric.} & \revgreen{Specialist orchestration overhead is \textbf{NR}.} & \codelink{https://github.com/simular-ai/Agent-S} \\
InfiGUIAgent \cite{liu2026infiguiagent} & EACL'26 & GUI/mm & compact reflection & \revgreen{Training uses \textbf{8 A800-80GB} GPUs and \textbf{32k} context.} & \revgreen{Exact inference latency and memory saving are \textbf{NR}.} & \na \\
\midrule
\tablegroup{System Efficiency Beyond the Planner}
ActionEngine \cite{zhong2026actionengine} & arXiv'26 & GUI/programs & runtime programs & \revgreen{Cost \textbf{\$0.71$\rightarrow$\$0.06} (\textbf{11.83$\times$}); latency \textbf{237.5$\rightarrow$118.3 s}; input \textbf{62.3k$\rightarrow$8.1k tokens}; calls \textbf{10.2$\rightarrow$1.8}.} & \revgreen{Compiler, validator, sandbox, and first-recovery overhead remain shifted costs.} & \na \\
CORE \cite{fan2025core} & NeurIPS'25 & Mobile/cloud & local-cloud routing & \revgreen{UI exposure reduced by \textbf{55.60\%}/\textbf{34.96\%}; latency \textbf{1.52--1.66$\times$} baseline; cloud tokens \textbf{0.94--1.15$\times$}.} & \revgreen{Privacy saving trades for local-routing latency.} & \na \\
GUIGuard \cite{wang2026guiguard} & arXiv'26 & GUI/privacy & privacy stages & \revgreen{\textbf{NR}; no privacy-detector runtime or deployment-overhead metric.} & \revgreen{Detector/runtime overhead is \textbf{NR}.} & \na \\
CoAct-1 \cite{song2025coact1} & ICLR'26 & OS/GUI+code & backend routing & \revgreen{Average steps: \textbf{10.15} vs. GUI baselines around \textbf{15} steps.} & \revgreen{Code backend overhead and safety cost are \textbf{NR}.} & \na \\
Agent S2 \cite{agashe2025agents2} & COLM'25 & OS/modular & split modules & \revgreen{\textbf{NR}; no module-routing latency, cost, or memory metric.} & \revgreen{Module orchestration overhead is \textbf{NR}.} & \codelink{https://github.com/simular-ai/Agent-S} \\
IntentCUA \cite{lee2026intentcua} & AAMAS'26 & OS/multi-agent & plan reuse & \revgreen{\textbf{NR}; public source PDF was unavailable in this extraction pass.} & \revgreen{Plan-reuse overhead is \textbf{NR}.} & \na \\
OS-Symphony \cite{yang2026ossymphony} & arXiv'26 & OS/framework & orch. runtime & \revgreen{\textbf{NR}; no message-protocol or orchestration-latency metric.} & \revgreen{Framework overhead is \textbf{NR}.} & \na \\
LongHorizonUI \cite{kang2026longhorizonui} & ICLR'26 & OS/reflection & rollback runtime & \revgreen{Average trajectory \textbf{24.6 steps}, max \textbf{37}; verifier choice changes latency \textbf{8.26 s$\rightarrow$5.74--6.59 s/step}.} & \revgreen{Rollback verifier-call overhead remains.} & \na \\
UltraCUA \cite{yang2025ultracua} & OpenRev'25 & GUI+prog. & GUI/API fusion & \revgreen{Reports \textbf{11\%} fewer steps overall; hybrid control uses \textbf{14.9\%} fewer steps.} & \revgreen{Tool/API routing latency and runtime cost are \textbf{NR}.} & \na \\
\end{longtable}
\endgroup

\end{document}